\PassOptionsToPackage{dvipsnames,table}{xcolor}
\documentclass[11pt,a4paper]{newtemplate}

\usepackage{amsmath,amsfonts,bm}

\def\eqref#1{equation~\ref{#1}}

\def\1{\bm{1}}

\DeclareMathAlphabet{\mathsfit}{\encodingdefault}{\sfdefault}{m}{sl}
\SetMathAlphabet{\mathsfit}{bold}{\encodingdefault}{\sfdefault}{bx}{n}

\title{\textbf{\wordmark}: Benchmarking Agents on Inverse Graphics of Dynamic Scenes}

\renewcommand\Affilfont{\normalfont\bfseries\small\centering}
\makeatletter\renewcommand\AB@affilsepx{\protect\\\protect\Affilfont}\makeatother
\author[1*\textdagger]{Ruihong Shen}
\author[2*]{\v{Z}iga Kova\v{c}i\v{c}}
\author[2,3*]{Peter Kulits}
\author[1]{Xingrui Wang}
\author[2]{Zizhang Li}
\author[4]{\authorcr Joshua B.~Tenenbaum}
\author[1]{Alan Yuille}
\author[2\textdaggerdbl]{Jieneng Chen}
\author[2\textdaggerdbl]{Jiajun Wu}
\affil[ ]{\mbox{\textsuperscript{1}Johns Hopkins University}\quad \mbox{\textsuperscript{2}Stanford University}\protect\\\mbox{\textsuperscript{3}Max Planck Institute for Intelligent Systems}\quad \mbox{\textsuperscript{4}Massachusetts Institute of Technology}}
\affil[ ]{\normalfont\small\href{https://4dcodebench.com}{4DCodeBench.com}\vspace{-42pt}}

\newcommand{\BENCHMARK}{\texorpdfstring{\textbf{\textsc{\wordmark}}}{4DCodeBench}\xspace}

\usepackage{natbib}
\usepackage{array}
\usepackage{subcaption}
\usepackage{wrapfig}
\newcommand{\fullpagefigure}[1]{\includegraphics[width=\textwidth,height=\dimexpr\textheight-3\baselineskip\relax,keepaspectratio]{#1}}
\newcommand{\halfpagefigure}[1]{\includegraphics[width=\textwidth,height=\dimexpr(\textheight-6\baselineskip)/2\relax,keepaspectratio]{#1}}
\usepackage[most]{tcolorbox}
\usepackage{fvextra}
\usepackage{titletoc}
\usepackage{xspace}

\newcommand{\cmark}{\textcolor{green!50!black}{\ding{51}}}
\newcommand{\xmark}{\textcolor{red}{\ding{55}}}

\usepackage{acro}
\DeclareAcronym{vlm}{short = VLM, long = vision-language model}

\newif\ifshowcomments
\showcommentstrue

\definecolor{myblue}{HTML}{3f55b5}
\definecolor{myorange}{HTML}{ec6b14}

\hypersetup{
    colorlinks=true,
    urlcolor=myorange,
    linkcolor=red,
    citecolor=myblue
}

\newcommand{\wordmark}{4DCodeBench}

\makeatletter
\newcommand{\authornotes}[1]{{\renewcommand\thefootnote{}\long\def\@makefntext##1{\leftskip\z@\rightskip\z@\parfillskip\z@ plus 1fil\noindent##1}\footnotetext{#1}}}
\makeatother

\fancypagestyle{firststyle}{
  \fancyhf{}
  \fancyhead[L]{{\small \textit{\the\month-\the\day-\the\year}}}
  \renewcommand{\footrulewidth}{0pt}
}

\teaser{%
  \authornotes{\textsuperscript{*}Equal contribution, listed in random order.\quad \textsuperscript{\textdaggerdbl}Equal co-advising.\quad \textsuperscript{\textdagger}Work done during a summer internship.}%
  \centering
  \includegraphics[width=\textwidth]{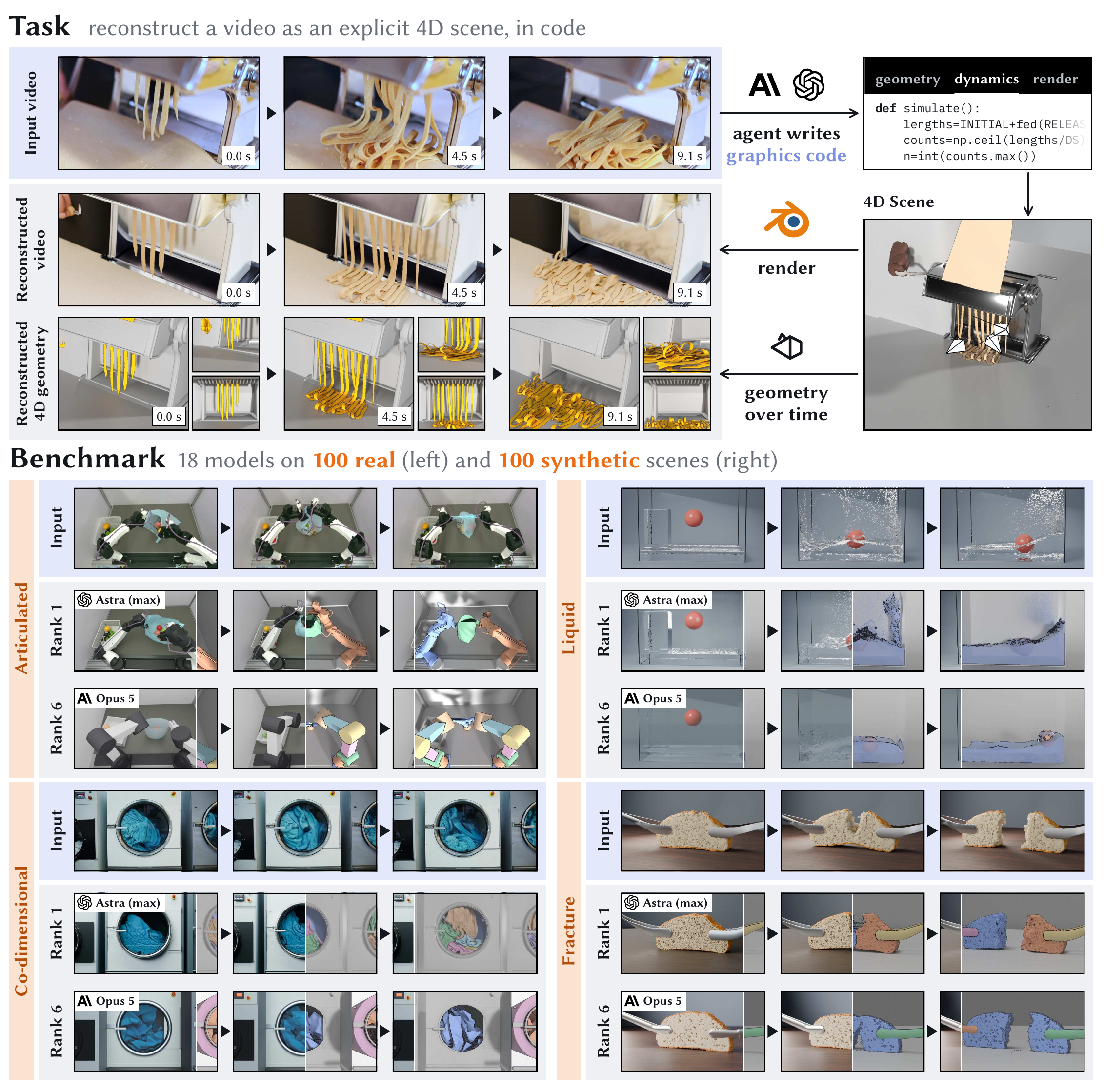}
  \captionof{figure}{\textbf{\BENCHMARK.} \textbf{Top:} the task. Given an input video, an agent must reconstruct it as an explicit 4D scene by writing graphics code; the scene is rendered back to video and its geometry is inspected over time. \textbf{Bottom:} we benchmark 18 models on 100 real and 100 synthetic scenes. Each example shows the input and two model reconstructions, split into render (left) and geometry (right).}
  \label{fig:teaser}
}

\begin{document}

\begin{abstract}
    
 We introduce \BENCHMARK, a benchmark for 4D inverse graphics through code generation, in which agents reconstruct dynamic scenes from video as executable graphics programs.
 To accomplish this, agents must translate visual observations into compact representations of scene structure and dynamics, by implementing abstractions such as physical simulations to reproduce complex behavior.
 To evaluate this capability, we curate a set of real-world videos and construct synthetic scenes spanning diverse physical phenomena, including deformation, fluid flow, and fracture.
 We perform extensive benchmarking of frontier models, finding that strong static reconstruction capabilities do not yet translate into reliable reconstruction of complex dynamics.
 \BENCHMARK provides a testbed for tracking progress toward agents that can interpret the dynamics of the world through code.
 Our benchmark is available at \href{https://github.com/4DCodeBench/4DCodeBench}{\nolinkurl{github.com/4DCodeBench/4DCodeBench}}. 
 
\end{abstract}

\maketitle

\section{Introduction}
\label{sec:intro}

Inverse graphics is often approached by fitting a prescribed model to visual observations.
Recovering an executable graphics program instead moves model specification into the inference problem.
The program specifies what geometry to construct and how to represent it.

Dynamic scenes render these modeling choices more consequential.
Flow, fracture, and deformation involve changes in shape and often topology, while material interactions affect each other's motion.
Representing such phenomena extends beyond the placement of objects or recovery from a moving camera in a static scene.
We ask how well multimodal coding agents can understand the dynamics of interacting objects and materials from video, choosing and implementing their own abstractions to reconstruct the scene.

We introduce \BENCHMARK to evaluate this capability through 4D inverse graphics.
Agents are given a reference video and tasked with writing code that generates the 3D geometry of the scene over time, and renders the result.
We leave implementation of these dynamics open rather than requiring the use of a particular approach.
This freedom renders the agents' choices themselves part of the investigation.

Our benchmark includes 100 real-world videos and 100 synthetic scenes that span diverse materials and interactions.
We separately evaluate appearance, geometry, and dynamics of the scene, comparing against reference videos and 3D ground truth on synthetic scenes.

Through evaluation of 18 frontier models, we observe that reconstruction of dynamics consistently lags behind that of appearance and static geometry.
While analytic motion is the most-common strategy, we find that leading models differ substantially in their approaches.

In summary, we contribute the following:
\begin{enumerate}
    \item \textbf{A benchmark and dataset for 4D inverse graphics through code generation.}
    We introduce \BENCHMARK, comprising 200 dynamic scenes -- 100 real videos and 100 synthetic scenes with 4D ground truth -- spanning diverse materials, interactions, and physical phenomena.
    Agents must reconstruct each video as an executable graphics program that exposes explicit geometry and dynamics over time.
    \item \textbf{A comprehensive evaluation of frontier and open-weight models.}
    We benchmark 18 multimodal coding models under a common execution environment and evaluation protocol, measuring visual fidelity, 2.5D and 3D geometry, and dynamics.
    Our analysis identifies substantial differences across models and shows that reconstructing dynamics remains a consistent weakness even when appearance and static geometry are recovered well.
    \item \textbf{Human-aligned evaluation and analysis of reconstruction failures.}
    We conduct a large-scale human preference study and develop automated VLM- and reconstruction-based metrics that closely track human judgments.
    Using these measurements, we analyze which scene properties and physical regimes expose different reconstruction failures, providing a diagnostic view of current model capabilities.
\end{enumerate}

\section{\BENCHMARK}
\label{sec:bench}

\begin{figure*}[t]
    \centering
    \includegraphics[width=\textwidth]{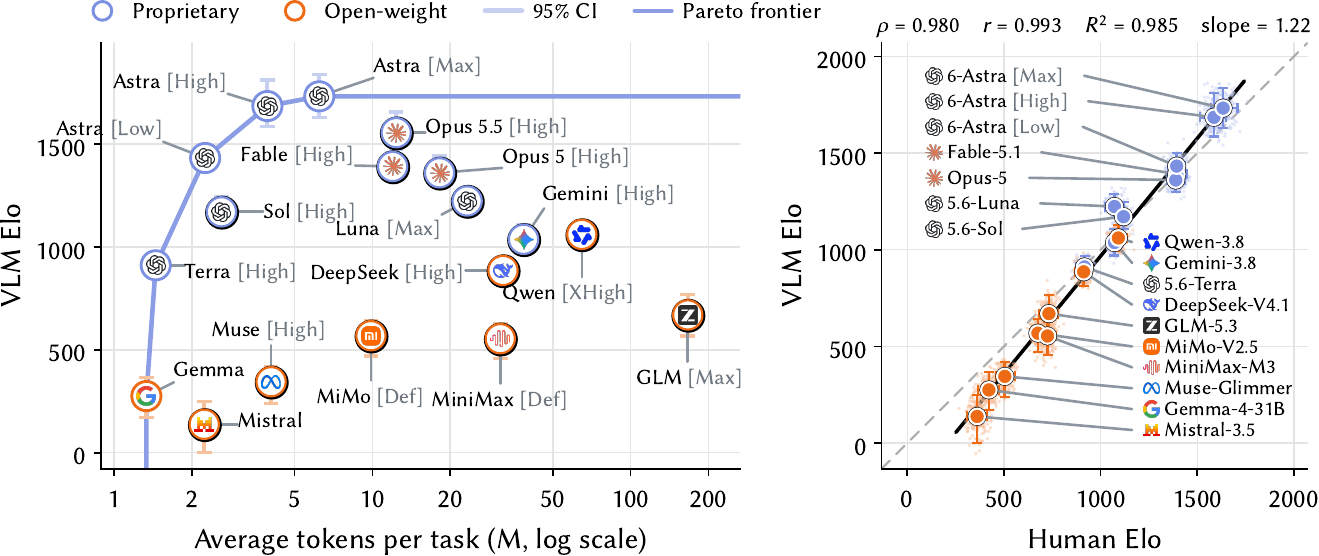}
\caption{\textbf{Cost--quality trade-off and human--\acs*{vlm} agreement.} The left panel plots each model's \acs*{vlm} Elo against its average token use per task, with the Pareto frontier; the right panel plots \acs*{vlm} Elo against human Elo for the 17 models in the human study. Error bars show 95\% intervals.}
\label{fig:elo}
\end{figure*}

\BENCHMARK comprises 200 scenes: 100 real-world videos curated from existing datasets and the web and 100 synthetic scenes built with physics simulations.
The two sources serve complementary roles.
Real-world videos capture realistic appearance and complex dynamics without the simplifying physical assumptions of a simulator, but lack 4D ground truth. Synthetic scenes depend on these assumptions but provide ground-truth geometry and motion at every frame, enabling direct evaluation in 3D and 4D space, rather than solely comparing rendered frames in image space. Below we describe the task definition (\S\ref{sec:def}), benchmark construction (\S\ref{sec:bench-construct}), data statistics (\S\ref{sec:dataset_stats}), and evaluation metrics (\S\ref{sec:metrics}).

\subsection{Task Definition: 4D Inverse Graphics Through Code Generation}
\label{sec:def}
Inverse graphics recovers the underlying representation of a scene from observations. We pose \emph{4D inverse graphics} as code generation: given a video, an agent writes an executable program -- which we term \emph{4D code} -- that constructs the scene's 3D geometry and its evolution over time.

Formally, given a reference video $x_{\mathrm{ref}}=(x_0,\ldots,x_{T-1})$ of a physical scene, an agent must produce executable graphics code $\Gamma=(\gamma_1,\ldots,\gamma_m)$, for 4D inverse graphics, constructing the scene's 3D geometry and describing its evolution over time. We impose no restrictions on the number, organization, or roles of the individual programs $\gamma_i$. Executing $\Gamma$ must expose a geometric state $s_\Gamma(t)$ at each evaluated timestep and render that state in Blender~\citep{blender} through a rendering procedure $\mathcal{R}_\Gamma$:
\begin{equation}
\Gamma=\mathcal{A}(x_{\mathrm{ref}}),
\qquad
\hat{x}_k=\mathcal{R}_{\Gamma}\left(s_{\Gamma}(k/r)\right),
\quad k=0,\ldots,T-1,
\end{equation}
where $\mathcal{A}(\cdot)$ denotes the agent and $r$ is the reference frame rate. $\mathcal{R}_\Gamma$ renders the geometric state using the camera, lighting, and materials specified by $\Gamma$. The rendered output must match the reference's resolution, frame count, and frame rate, and $\Gamma$ must execute without access to $x_{\mathrm{ref}}$. We render in Blender, but the code implementation is simulator-agnostic: dynamics may be keyframed or simulated with any tool such as Taichi~\citep{hu2021quantaichi} or Warp~\citep{Macklin_Warp_A_High-performance_2022}. Figure~\ref{fig:supp:task} gives an intuitive overview of our task pipeline, which supports diverse agents and videos.

\subsection{Benchmark Overview and Construction}
\label{sec:bench-construct}
To construct the benchmark, we collect real-world videos and generate synthetic scenes covering a wide range of physical dynamics. These include scenes of robotic manipulation, controlled physics demonstrations, and everyday activities. We organize scenes into four primary categories: rigid and articulated bodies, deformable solids, co-dimensional structures such as cloth and rope, and flowing materials such as grains and fluids. Across these categories, scenes exhibit phenomena including collisions, deformation, flow, and fracture. We further characterize scenes by object count, material composition, and dynamics regime, distinguishing passive dynamics from externally driven motion, such as robotic manipulation or prescribed motion of rigid objects.

\noindent\textbf{Real-world scenes.}
We curate videos from existing physical-understanding, video-generation, and robotics datasets, supplemented with web-sourced footage. We select clips with diverse physical behaviors while minimizing repeated scenarios, prioritizing stationary cameras, limited occlusion, and continuous footage without cuts or editing-induced discontinuities. We also exclude humans and animals and minimize visible hands, since reconstructing their detailed geometry and articulation introduces challenges beyond our focus on dynamics. Each clip is manually selected and reviewed for visual quality and temporal continuity. Appendix~\ref{sec:supp:real_data} lists the video sources and the number of scenes from each source.

\noindent\textbf{Synthetic scenes.}
We construct synthetic scenes using existing physical simulators, extending their implementations where needed to support additional material models. Scenes are designed by the authors, combining procedurally generated geometry with publicly available meshes such as the Stanford bunny and armadillo.  We select simulators suited to each scene's desired materials and interactions and use them to compute the objects' geometry and motion over time. We design diverse scenarios that vary geometry, material types and parameters, initial conditions, and rendering settings to create varied reconstruction tasks. Authors with experience in physical simulation developed and reviewed each scene for numerical instability, unintended interpenetration, and other visible simulation artifacts. Appendix~\ref{sec:supp:synthetic} and Table~\ref{tab:supp:simulators} list the simulators used and provides a breakdown of scenes by simulator.

\subsection{Dataset Statistics}
\label{sec:dataset_stats}
\begin{figure*}[t]
    \centering
    \includegraphics[width=\textwidth]{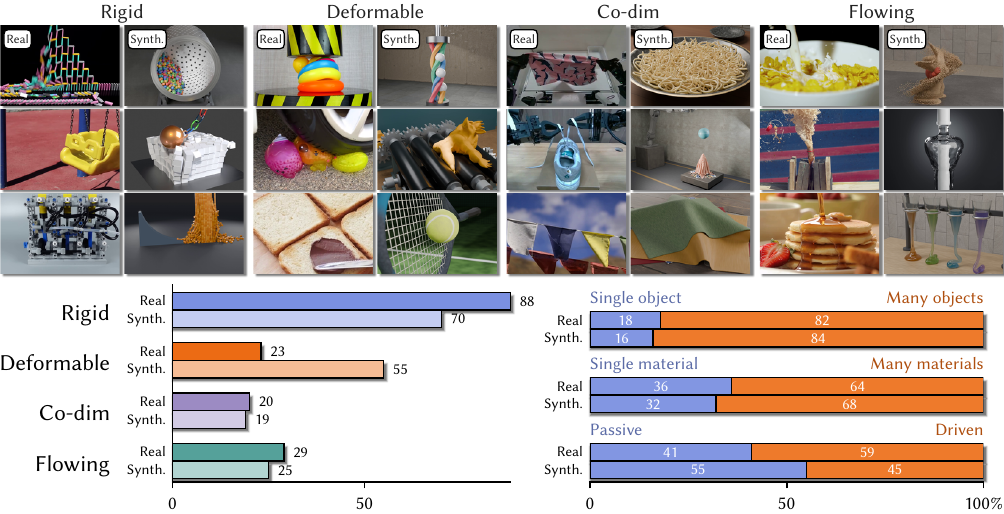}
    \vspace{-0.5cm}
    \caption{
    \textbf{Dataset examples and statistics.} The top row shows a real and a synthetic example scene for each matter family. The bottom panels count the real and synthetic scenes that contain each family (left) and break both splits down by the number of dynamic objects, the number of materials, and whether the motion is passive or driven (right).
    }
    \label{fig:dataset_statistics}
    \vspace{-1em}
\end{figure*}

Figure~\ref{fig:dataset_statistics} summarizes the distribution of scene properties in \BENCHMARK. We characterize each scene by the physical behavior of its primary dynamic objects, together with properties such as the number of dynamic objects, the presence of multiple material types, and whether the motion is passive or externally driven.

\noindent\textbf{Scene ontology.}
We organize scenes hierarchically by the matter of their main dynamic objects. We distinguish four broad classes: \emph{rigid and articulated}, including rigid bodies and articulated systems; \emph{deformable solids}, including hyperelastic, viscoelastic, and elastoplastic materials; \emph{co-dimensional} structures, including shells and rods; and \emph{flowing} materials, including granular media, Newtonian fluids, and viscoplastic fluids. Scenes may contain multiple matter types interacting with each other. The full breakdown is visualized in Appendix~\ref{sec:supp:dataset}.

Beyond matter type, we characterize scenes by the number of dynamic objects and by their interactions. \textbf{83\%} of scenes contain multiple dynamic objects, and \textbf{66\%} contain multiple materials. These include interactions such as rigid--deformable, rigid--fluid, fluid--deformable, deformable--co-dimensional, and rigid--co-dimensional coupling. We also distinguish passive dynamics, where motion follows from the initial state and subsequent physical interactions, from externally driven dynamics, such as robotic manipulation or prescribed motion of an object. The distribution of these properties across our scenes, together with the breakdown across the real and synthetic splits, is shown in Figure~\ref{fig:dataset_statistics}.

All videos are capped at a maximum frame rate of 60\,fps during dataset construction, and real videos are capped at 300 frames. Videos with a higher source frame rate are temporally subsampled to at most 60\,fps, while videos recorded at lower frame rates retain their original frame rate. Figure~\ref{fig:supp:dataset_distributions} reports the final distributions of video length and frame rate. 

\subsection{Metrics and Evaluation: What Is a Good Reconstruction?}
\label{sec:metrics}
A faithful 4D reconstruction must reproduce both the visual appearance and the physical dynamics of the observed scene.
Because ground-truth availability is asymmetric -- synthetic scenes provide full 4D world states while real videos provide only monocular views -- our evaluation spans five metric families: \textbf{Perceptual}, \textbf{2D Dynamics}, and \textbf{2.5D Geometry} measured against the reference video, alongside \textbf{3D Geometry} and \textbf{3D Dynamics} evaluated against the reference world.
The \textbf{Overall} score averages these five families; failed runs receive the worst score on the affected metrics (Appendix~\ref{sec:supp:executability}).

\noindent\textbf{Video-reference evaluations.}
On all scenes, Perceptual measures visual and semantic fidelity to the reference video using DINOv3~\citep{simeoni2025dinov3} similarity, and 2D Dynamics measures the coverage of moving objects (\emph{Dynamic IoU}). On real videos, we additionally estimate optical flow, point trajectories, and depth from the reference video with off-the-shelf vision models; 2D Dynamics compares the flow (\emph{Flow}) and trajectories (\emph{Track2D}), and 2.5D Geometry compares the depth (\emph{Depth error}). On the reconstruction side, these quantities are computed analytically from the 4D world rather than estimated from rendered pixels (Appendices~\ref{sec:supp:metrics:all} and~\ref{sec:supp:metrics:real}).

\noindent{\textbf{\acs*{vlm}-based VQA and pairwise evaluation.}}
We further evaluate reconstructions with a \ac{vlm} as a judge. VQA asks scene-specific questions in four categories: initial state, final state, key events, and contact; questions are gated by the presence of the scene's main object (Figure~\ref{fig:vqa_overall}; Appendix~\ref{sec:supp:vqa}). 
Pairwise evaluation presents the reference video together with two reconstructions and asks which more closely matches the reference in geometry and dynamics. Pairwise preferences are aggregated across scenes into a model-level Elo rating using a Bradley--Terry model (Appendix~\ref{sec:vlm_pairwise}).

\noindent{\textbf{4D-reference evaluations.}}
On synthetic scenes, after aligning the reconstruction to the reference (Appendix~\ref{sec:supp:metrics:synthetic}), 3D Geometry measures static surface accuracy at frame~0 via Chamfer distance (\emph{Scene 3D}). 3D Dynamics compares motion from two complementary views: a Lagrangian perspective tracking matter along persistent 3D paths (\emph{Trajectory DTW}), and an Eulerian perspective comparing frame-to-frame displacement distributions without correspondence (\emph{EMD step}).

\noindent{\textbf{Geometry diagnostics.}}
We also check each delivered 4D world for structural defects: open boundaries, non-manifold edges, degenerate faces, self-intersections, and object interpenetrations (Figure~\ref{fig:supp:geometry_defects}). Because these checks assess mesh validity rather than fidelity to the reference, they are reported as diagnostics outside the Overall score (Appendix~\ref{sec:supp:additional_metrics}).

\section{Experiments}
\label{sec:experiments}

\subsection{Experimental Setup}
\label{sec:experimental_setup}
\newcommand{\inlinelogo}[1]{\raisebox{-0.12em}{\includegraphics[height=0.85em,width=1.1em,keepaspectratio]{figures/logos/#1.pdf}}\,}
We evaluate \textbf{18} multimodal coding models, covering both proprietary models (\inlinelogo{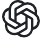}GPT-6 Astra [Max, High, Low], \inlinelogo{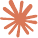}Claude Fable 5.1 [High], \inlinelogo{claude}Claude Opus 5 [High], \inlinelogo{claude}Claude Opus 5.5 [High], \inlinelogo{gpt}GPT-5.6 Sol [High], \inlinelogo{gpt}GPT-5.6 Luna [Max], \inlinelogo{gpt}GPT-5.6 Terra [High], \inlinelogo{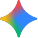}Gemini 3.8 Flash [High]) and open-weight models (\inlinelogo{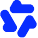}Qwen3.8 Flash-Next [XHigh], \inlinelogo{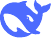}DeepSeek v4.1 Flash [High], \inlinelogo{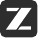}GLM 5.3 Flash [Max], \inlinelogo{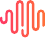}MiniMax M3 [Def], \inlinelogo{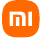}MiMo v2.5 [Def], \inlinelogo{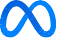}Muse Glimmer [High], \inlinelogo{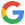}Gemma-4 31B, \inlinelogo{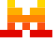}Mistral Medium 3.5).\footnote{We also evaluated Kimi K3 but omit it from the reported results: its unusually long agent trajectories cost about 50\% more than Fable, without a commensurate gain in reconstruction quality.} We evaluate multiple reasoning-effort settings of Astra to study the effect of inference-time reasoning on reconstruction performance. We run each model once on each of the 200 benchmark scenes. 

Each benchmarked agent takes as input only the reference RGB video and a fixed task prompt describing the reconstruction task and required output format. We do not provide scene names, semantic descriptions, object lists, or any other information about the scene; agents must infer these from the input video. The complete prompt is provided in Appendix~\ref{sec:supp:agent_prompt}.

Proprietary models run in their provider's native agent CLI, and open-weight models in the open Stirrup harness~\citep{stirrup}. Each run executes in an isolated container with one GPU, an empty workspace, and read-only access to the reference video and task files. The environment provides Blender, common numerical and simulation libraries, and an offline copy of the Blender API documentation, so agents may implement dynamics with Blender, the provided libraries, or their own code. Appendix~\ref{sec:supp:harness} details the environment.

Each agent must return an executable program together with a reconstructed scene containing the rendered video, camera parameters, per-frame geometry, and temporal correspondence for dynamic matter. The program must regenerate the reconstruction from scratch without accessing the reference video. We describe the exact file structure, coordinate conventions, and validation requirements in Appendix~\ref{sec:supp:output_format}. Note that we do not restrict how dynamics are implemented; for example, animation and physical simulation are two possible approaches. Dynamics may be implemented in Blender, using simulation libraries such as Taichi~\citep{hu2019taichi,hu2019difftaichi,hu2021quantaichi} or Warp~\citep{Macklin_Warp_A_High-performance_2022}, or with custom code. The program must also expose its geometric state in each evaluated frame, allowing evaluation of both rendered appearance and, for synthetic scenes, geometry and motion against 4D ground truth. 

Each model is evaluated on all 200 benchmark scenes. If a run fails to produce a valid rendered video or world representation, metrics requiring the missing output are assigned their worst value (Section~\ref{sec:metrics}). Submission success rates are reported in Appendix~\ref{sec:supp:executability}.

\subsection{Human Evaluation}
\label{sec:human_eval}
We run a pairwise human preference study on all 200 scenes to measure perceived reconstruction quality and to validate the automated metrics. Each trial shows the reference video, then the renders of two models side by side in random order; participants pick the one closer to the reference in geometry, layout, and dynamics, or mark them \emph{about the same} (10.5\% of responses). Model pairs are sampled adaptively to shrink the widest confidence intervals~\citep{chiang2024chatbot}, and judgments are aggregated into Elo ratings with a Bradley--Terry model~\citep{bradley1952rank} as for the \ac{vlm} judge. The study covers 17 of the 18 models (all but Opus 5.5 [High]). We collect 3,587 judgments from 76 participants; 209 comparisons are judged by several participants and 916 also by the \ac{vlm}, which lets us measure inter-rater and human--\ac{vlm} agreement (Section~\ref{sec:overall_results}; Appendix~\ref{sec:supp:human_study}).

\section{Results and Ablations}
\label{sec:results}

\begin{figure*}[t]
    \centering
    \includegraphics[width=\textwidth]{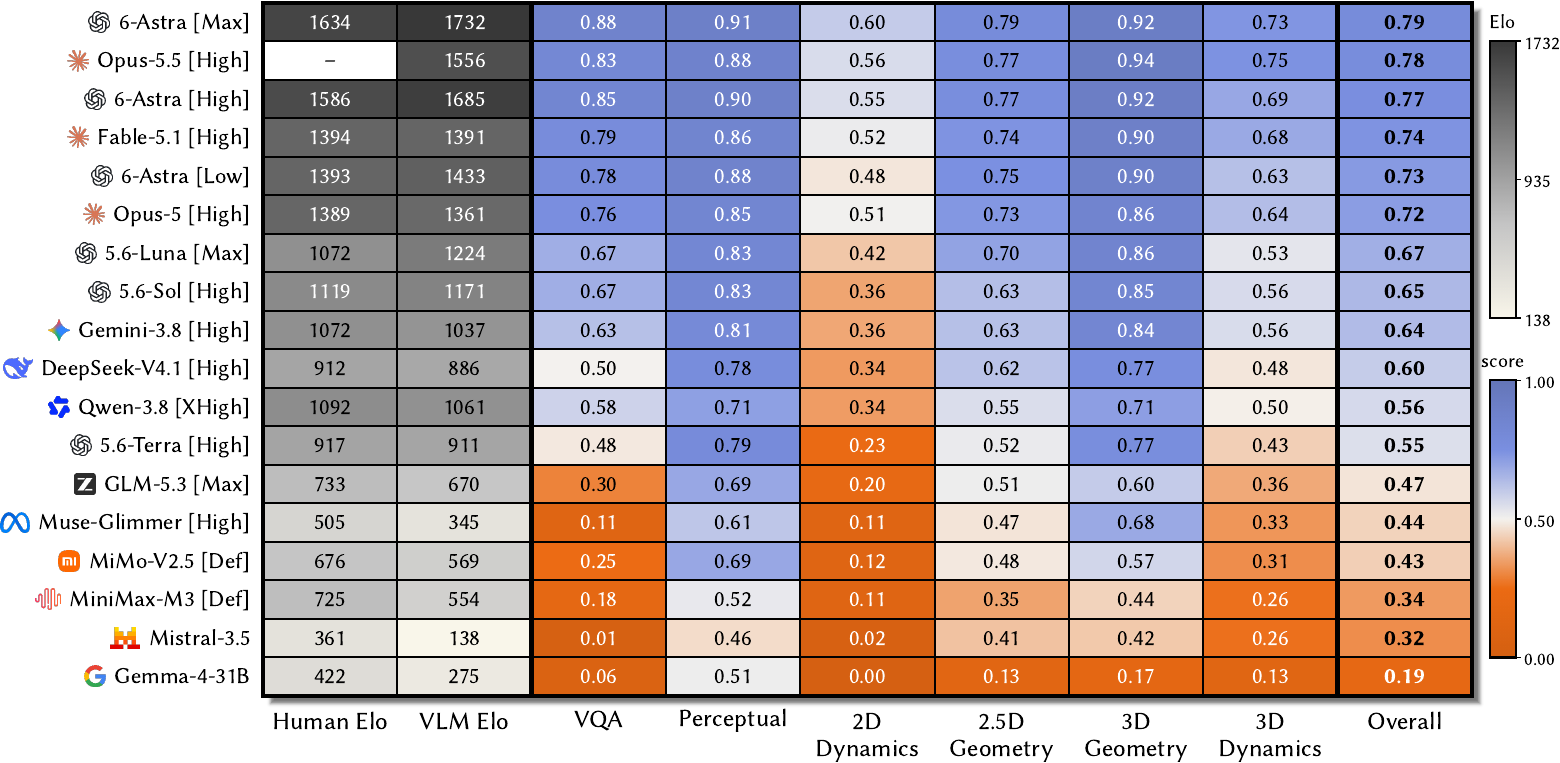}
\caption{\textbf{Leaderboard of 18 models on 200 scenes.} Each row lists the human and \acs*{vlm} Elo, VQA accuracy, the five family scores in $[0,1]$ (higher is better), and Overall, the mean of Perceptual, 2D Dynamics, 2.5D Geometry, 3D Geometry, and 3D Dynamics; rows are sorted by Overall. Opus 5.5 [High] was not part of the human study due to its time of release (--).}
\label{fig:main_leaderboard}
\vspace{-0.15cm}
\end{figure*}

\subsection{Overall Performance and Human Alignment}
\label{sec:overall_results}

\begin{figure*}[t]
    \centering
    \includegraphics[width=\textwidth]{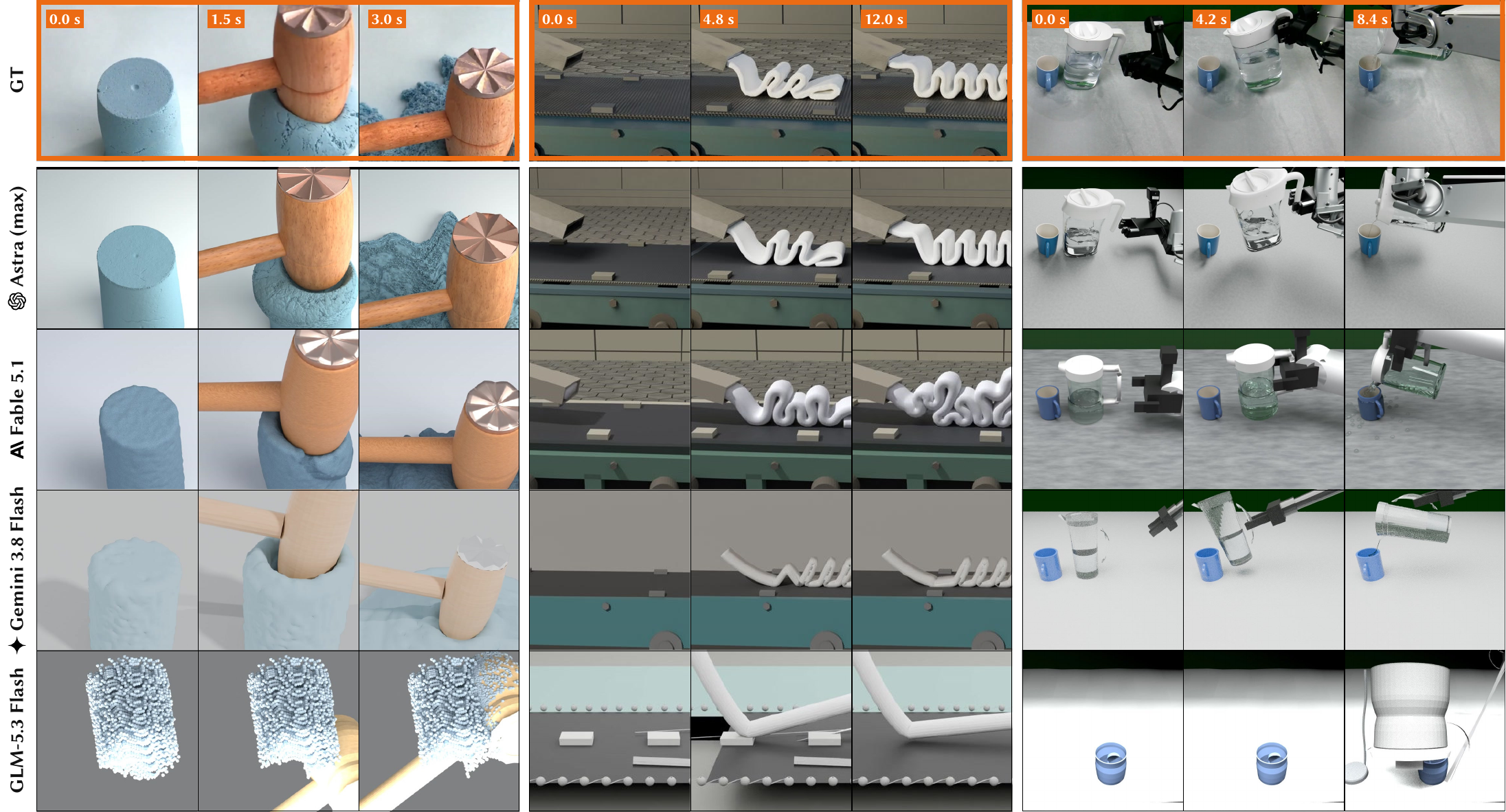}
\caption{\textbf{Qualitative comparison across models.} Each row shows one model's reconstruction (top: ground truth; then: GPT-6 Astra [Max], Claude Fable 5.1, Gemini 3.8 Flash, and GLM 5.3 Flash). Stronger models recover both geometry and dynamics better.}
\label{fig:model_comparison}
\vspace{-1em}
\end{figure*}

We evaluate 18 models on 200 scenes (Figure~\ref{fig:main_leaderboard}). Astra [Max] leads the Overall ranking, followed by Opus 5.5 [High], Astra [High], Fable [High], and Astra [Low]. Among the models evaluated, open-weight models generally trail behind proprietary ones. The leaderboard distinguishes substantial differences in reconstruction quality, including between a wide range of open-weight models. Failed submissions receive the worst applicable metric scores; 90.1\% of submissions are fully executable (Figure~\ref{fig:executability}). Figures~\ref{fig:model_comparison} and~\ref{fig:teaser} show reconstruction examples across several models. More sample results are included in Appendix~\ref{sec:supp:additional_results} and on the \href{https://4dcodebench.com}{project page}.

Human judgments support this ranking. Across 3,587 pairwise judgments from 76 participants, human and VLM Elo correlate very strongly across models (Spearman $\rho=0.980$; Figure~\ref{fig:elo}). We additionally measure agreement on the shared set of individual comparisons. On 916 shared comparisons, human and VLM judgments agree in 89.3\% of cases ($\kappa=0.761$), near the 92.1\% agreement between human annotators on repeated comparisons ($\kappa=0.816$). Human Elo also correlates with the Overall score ($\rho=0.96$).
Individual \ac{vlm} judgments are less reliable for closely matched models: agreement is near chance within 50 Elo points, reaching 75\% at a 132-point separation and 90\% at 343 points. 

\begin{figure*}[t]
    \centering
    \includegraphics[width=\textwidth]{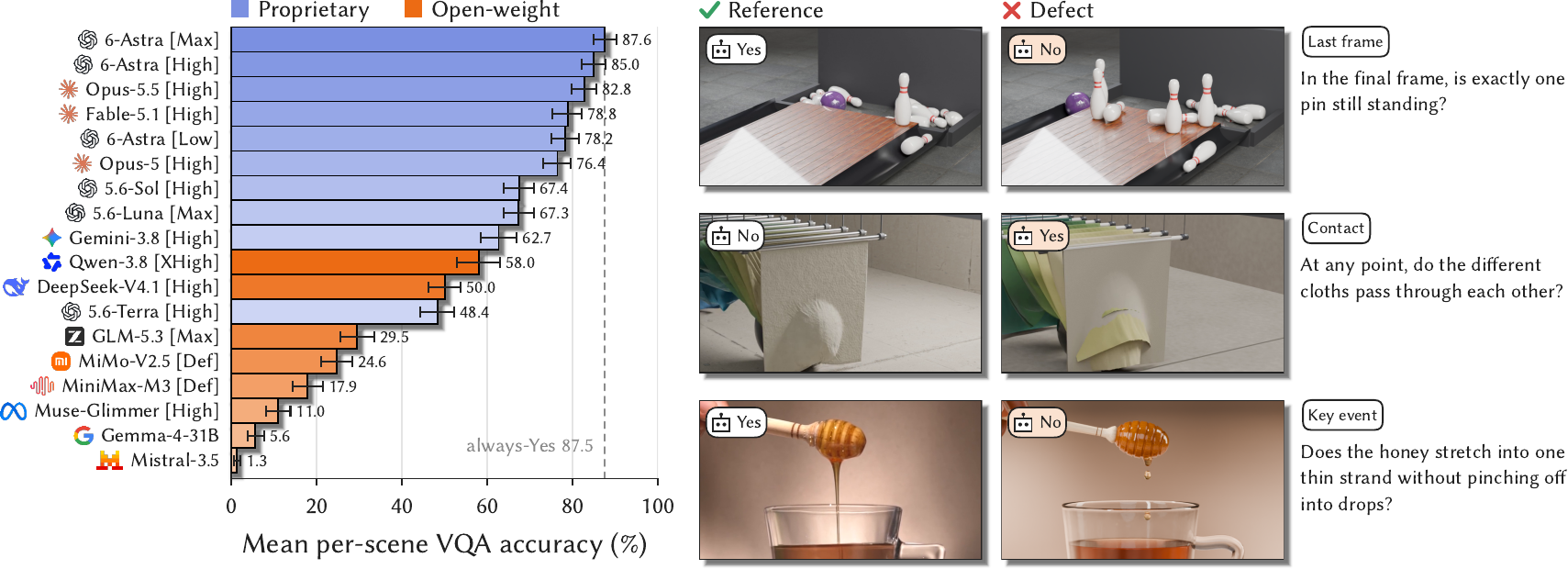}
    \caption{\textbf{Visual question answering (VQA).} The left panel reports each model's mean per-scene accuracy; the dashed line is the accuracy of always answering Yes. The right panel illustrates the protocol with example questions, each asked on the reference video and on an edited render that breaks the queried property.}
    \label{fig:vqa_overall}
\end{figure*}

VQA provides a complementary assessment of individual reconstructions (Figure~\ref{fig:vqa_overall}). Astra [Max] reaches a mean per-scene accuracy of 87.6\%, compared with 78.8\% for Fable [High] and 29.5\% for GLM [Max]. These averages also conceal large differences across scenes: Astra [Max] answers every question on 59\% of scenes, whereas Mistral answers none on 94\% (Figure~\ref{fig:supp:vqa_scene_correctness}).

Figure~\ref{fig:elo} (left) shows that greater token use across different models does not consistently yield higher Elo. The same pattern appears in the running statistics: token volume and agent steps have little correlation with Overall score across models (Figure~\ref{fig:supp:running_stats}).

Together, the agreement with human preferences supports using the automated evaluation suite to compare new models on \BENCHMARK, while the separate scores identify where their reconstructions succeed or fail.

\subsection{Performance Across Metric Families}
\label{sec:metric_results}

\begin{figure*}[t]
    \centering
    \begin{minipage}[b]{0.485\textwidth}
        \centering
        \includegraphics[width=\linewidth]{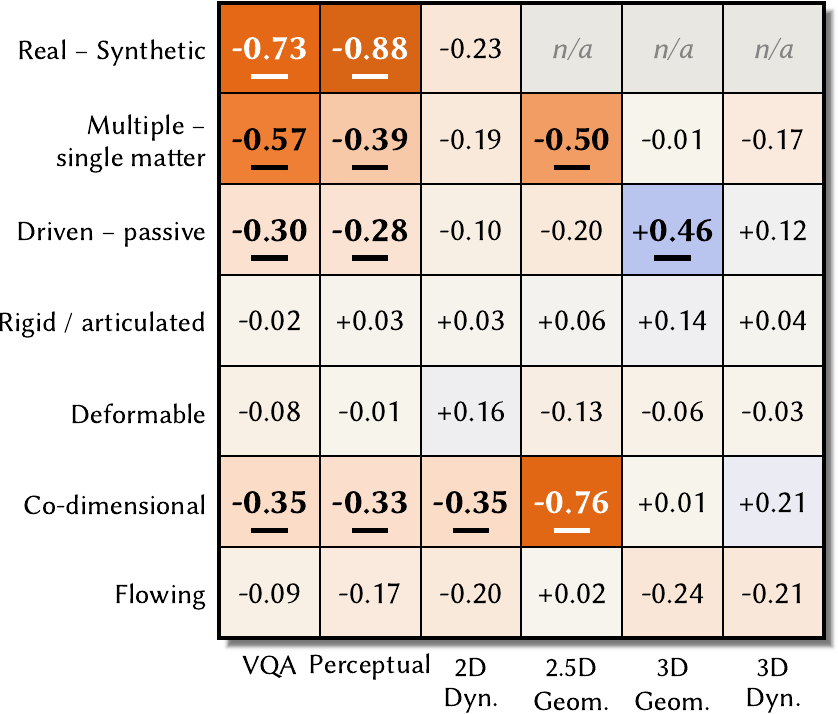}
    \end{minipage}\hfill
    \begin{minipage}[b]{0.485\textwidth}
        \centering
        \includegraphics[width=\linewidth]{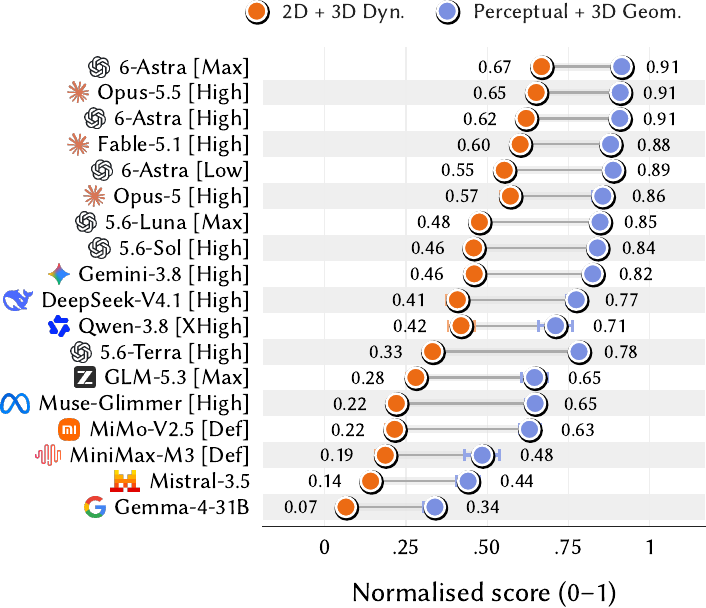}
    \end{minipage}\par
    \vspace{-0.1cm}
    \begin{minipage}[t]{0.485\textwidth}
        \caption{\textbf{Category effects.} Each cell is the standardized change in a score on one category of scenes, measured against its counterpart or, for a matter family, against all scenes; underlined effects have a 95\% interval that excludes zero and the same sign for at least 14 of 18 models.}
        \label{fig:main_category}
    \end{minipage}\hfill
    \begin{minipage}[t]{0.485\textwidth}
        \caption{\textbf{Static--dynamic gap.} For each model, the blue dot averages the static families (Perceptual, 3D Geometry) and the orange dot the dynamic ones (2D and 3D Dynamics), with 95\% intervals over scenes.}
        \label{fig:dynamic_static_gap}
    \end{minipage}\par
    \vspace{-0.6cm}
\end{figure*}

The leaderboard reveals a consistent weakness in reconstructing motion (Figure~\ref{fig:main_leaderboard}). Across all 18 models, Perceptual and 3D Geometry scores are higher on average than 2D and 3D Dynamics scores (Figure~\ref{fig:dynamic_static_gap}). The gap remains when appearance is compared with 2D motion and 3D geometry with 3D motion separately (Figure~\ref{fig:supp:family_gaps}). Even the strongest model, Astra [Max], scores 0.91 on the static families versus 0.67 on the dynamic families. Reconstructing a scene's evolution remains harder than recovering its appearance and geometry. 

The metrics give broadly similar rankings across models (Figure~\ref{fig:main_leaderboard}), but agree less when ranking reconstructions of the same scene (Figure~\ref{fig:supp:metric_concordance}). They therefore expose differences in how well each model reconstructs appearance, geometry, and motion on individual scenes. VQA shows a related temporal distinction: first-frame questions are generally answered more reliably than questions about the final frame and key events (Appendix Figure~\ref{fig:supp:vqa_categories}).

Five additional reconstruction metrics broadly agree with the leaderboard, whereas mesh-quality checks do not: simple but inaccurate geometry can score well on watertightness and related measures (Figure~\ref{fig:supp:additional_metrics}). We therefore report mesh-quality checks separately as diagnostics of structural defects.

\subsection{Performance Across Scene Types}
\label{sec:ontology_results}

We next examine how reconstruction quality varies with the scene ontology of Section~\ref{sec:bench}. Figure~\ref{fig:main_category} reports standardized differences across scene categories. Relative to synthetic scenes, real scenes have lower VQA and Perceptual scores ($-0.73$ and $-0.88$). Scenes containing multiple matter types show a similar pattern relative to single-matter scenes ($-0.57$ and $-0.39$). Thus, both the source of a scene and its material composition affect its recognizable appearance and events. 

Other categories expose different weaknesses. Co-dimensional scenes have lower 2D Dynamics and 2.5D Geometry scores ($-0.35$ and $-0.76$), while driven scenes have lower VQA but higher 3D Geometry scores ($-0.30$ and $+0.46$). The highlighted effects recur across most individual models, rather than arising only from pooled averages. Flowing scenes also show lower 3D Dynamics ($-0.21$), although this effect does not meet the figure's consistency criterion. Together, these patterns show why scene categories must be considered alongside aggregate model scores.

\subsection{How Do Agents Model Dynamics?}
\label{sec:code_analysis}
The executable submissions let us examine how agents implement motion. Table~\ref{tab:solution-strategies} classifies their solutions as analytic motion, custom simulation, Blender physics, or keyframing. Overall, 67\% of solutions use analytic motion, 19\% custom simulation, 10\% Blender physics, and 3\% keyframing. The distribution varies substantially across models: Opus 5.5 [High] uses custom simulation in 61\% of its solutions, while Astra [Low] uses analytic motion in 85\% and Terra [High] in 100\%. Models use different representations of dynamics even when solving the same reconstruction task. Appendix~\ref{supp:describe_motion} describes what each category contains in more detail.

\subsection{Effect of Inference-Time Reasoning}
\label{sec:reasoning_ablation}
Increasing Astra's reasoning effort from Low to High to Max raises its Overall score from 0.73 to 0.77 to 0.79 (Figure~\ref{fig:main_leaderboard}). Its 2D Dynamics score increases from 0.48 to 0.55 to 0.60, and its 3D Dynamics score from 0.63 to 0.69 to 0.73. VQA similarly increases from 78.2\% to 85.0\% to 87.6\%. These gains accompany greater output-token use and cost (Figure~\ref{fig:supp:running_stats}). Unlike the weak relationship between token use and quality across different models, additional reasoning within Astra consistently improves its reconstruction scores.

\section{Related Work}
\label{sec:related}
Recent benchmarks have begun evaluating coding agents on spatial and physical tasks, from constructing 3D objects to implementing simple simulations. 3DCodeBench tests whether agents can model 3D objects in code from text or images~\citep{gao20263dcodebench}. PhysCodeBench tests whether they can write executable physics simulations from text descriptions~\citep{xie2026physcodebench}. 

VisPhyWorld instead gives agents synthetic physics videos to reconstruct as simulator code~\citep{liang2026visphyworld}. MPMWorlds also reconstructs synthetic videos, covering deformation, material changes, and topology-changing events in 2D~\citep{kovavcivc2026mpmworlds}. BVB reconstructs primarily static indoor scenes observed by a moving camera~\citep{tang2026bvb}. \BENCHMARK instead curates videos with minimal camera motion to test whether agents reconstruct changes in scene geometry and physical motion, which we also evaluate directly against 3D ground truth on synthetic scenes. See Appendix~\ref{sec:supp:related_work} for further review and discussion.

\section{Conclusion}
\label{sec:conclusion}

\noindent\textbf{Limitations.} Our current evaluation focuses on reconstruction fidelity, leaving the underlying physical mechanisms underconstrained. Prescribed trajectories can achieve high scores, while the generalization benefits of simulation remain largely untested. Interventions on initial conditions and external forces, together with longer-horizon prediction, would be a good way to assess transferable physical understanding in future work. Our evaluation also focuses on final submissions; tracking intermediate reconstructions across render-and-compare rounds would reveal how quality improves with iteration and token expenditure, when gains saturate, and how effectively agents use visual feedback.

\noindent\textbf{Discussion.} \BENCHMARK highlights several directions for improving agents that reconstruct dynamic scenes through graphics code. The persistent gap between static and dynamic reconstruction underscores the challenge of inferring material properties, interactions, and driving forces from video and translating them into accurate physics solvers. Leading models adopt markedly different reconstruction strategies, ranging from predominantly analytic motion to custom physical simulations. This raises the question of when simulation improves reconstruction quality and whether requiring physics-based simulation would preserve the relative performance of these models. Meanwhile, greater token expenditure across models does not consistently yield better reconstructions, highlighting the importance of reasoning efficiency. Stronger physical intuition could help agents form plausible hypotheses earlier, anticipate the consequences of modeling choices, and use targeted feedback to refine their solutions with fewer render-and-compare iterations.

We introduced \BENCHMARK to evaluate agents' ability to reconstruct videos as executable 4D scenes.
Across the models studied, automated rankings closely agree with human preferences.
While the strongest models excel at reconstruction of appearance and geometry, a clear gap remains in their ability to abstract dynamics.
By measuring these capabilities across diverse interactions, \BENCHMARK provides a way to track progress in 4D inverse graphics and identify where models still fail.

\newpage
{
\small
\bibliographystyle{iclr2027_conference}
\bibliography{reference_header,main}

\begin{thebibliography}{79}
\providecommand{\natexlab}[1]{#1}
\providecommand{\url}[1]{\texttt{#1}}
\expandafter\ifx\csname urlstyle\endcsname\relax
  \providecommand{\doi}[1]{doi: #1}\else
  \providecommand{\doi}{doi: \begingroup \urlstyle{rm}\Url}\fi

\bibitem[Allshire et~al.(2026)Allshire, Singh, Singh, Rashid, Choi, McAllister, Yu, Chen, Huang, Abbeel, et~al.]{allshire2026scalable}
Arthur Allshire, Himanshu~Gaurav Singh, Ritvik Singh, Adam Rashid, Hongsuk Choi, David McAllister, Justin Yu, Yiyuan Chen, Huang Huang, Pieter Abbeel, et~al.
\newblock Scalable behavior cloning with open data, training, and evaluation.
\newblock \emph{arXiv preprint arXiv:2606.27375}, 2026.

\bibitem[Ando(2024)]{ando2024cubic}
Ryoichi Ando.
\newblock A cubic barrier with elasticity-inclusive dynamic stiffness.
\newblock \emph{ACM Transactions on Graphics (TOG)}, 2024.

\bibitem[{Artificial Analysis}(2025)]{stirrup}
{Artificial Analysis}.
\newblock Stirrup: The lightweight framework for building agents.
\newblock \url{https://github.com/ArtificialAnalysis/Stirrup}, 2025.

\bibitem[Bansal et~al.(2025)Bansal, Lin, Xie, Zong, Yarom, Bitton, Jiang, Sun, Chang, and Grover]{bansal2025videophy}
Hritik Bansal, Zongyu Lin, Tianyi Xie, Zeshun Zong, Michal Yarom, Yonatan Bitton, Chenfanfu Jiang, Yizhou Sun, Kai-Wei Chang, and Aditya Grover.
\newblock {VideoPhy}: Evaluating physical commonsense for video generation.
\newblock In \emph{{Proceedings of International Conference on Learning Representations (ICLR)}}, 2025.

\bibitem[Battaglia et~al.(2013)Battaglia, Hamrick, and Tenenbaum]{battaglia2013simulation}
Peter~W. Battaglia, Jessica~B. Hamrick, and Joshua~B. Tenenbaum.
\newblock Simulation as an engine of physical scene understanding.
\newblock \emph{PNAS}, 2013.

\bibitem[Bender et~al.(2026)]{SPlisHSPlasH_Library}
Jan Bender et~al.
\newblock {SPlisHSPlasH Library}, 2026.
\newblock URL \url{https://github.com/InteractiveComputerGraphics/SPlisHSPlasH}.

\bibitem[Besl \& McKay(1992)Besl and McKay]{besl1992method}
Paul~J. Besl and Neil~D. McKay.
\newblock A method for registration of {3-D} shapes.
\newblock \emph{{Transactions on Pattern Analysis and Machine Intelligence (TPAMI)}}, 14\penalty0 (2):\penalty0 239--256, 1992.
\newblock \doi{10.1109/34.121791}.

\bibitem[{Blender Foundation}(2026)]{blender}
{Blender Foundation}.
\newblock Blender.
\newblock \url{https://www.blender.org}, 2026.

\bibitem[Bordes et~al.(2025)Bordes, Garrido, Kao, Williams, Rabbat, and Dupoux]{bordes2025intphys}
Florian Bordes, Quentin Garrido, Justine~T. Kao, Adina Williams, Michael Rabbat, and Emmanuel Dupoux.
\newblock {IntPhys} 2: Benchmarking intuitive physics understanding in complex synthetic environments.
\newblock \emph{arXiv preprint arXiv:2506.09849}, 2025.

\bibitem[Bradley \& Terry(1952)Bradley and Terry]{bradley1952rank}
Ralph~Allan Bradley and Milton~E. Terry.
\newblock Rank analysis of incomplete block designs: {I}. {T}he method of paired comparisons.
\newblock \emph{Biometrika}, 39\penalty0 (3/4):\penalty0 324--345, 1952.

\bibitem[Bu et~al.(2025)Bu, Cai, Chen, Cui, Ding, Feng, Gao, He, Hu, Huang, et~al.]{bu2025agibot}
Qingwen Bu, Jisong Cai, Li~Chen, Xiuqi Cui, Yan Ding, Siyuan Feng, Shenyuan Gao, Xindong He, Xuan Hu, Xu~Huang, et~al.
\newblock {AgiBot World Colosseo}: A large-scale manipulation platform for scalable and intelligent embodied systems.
\newblock In \emph{{IEEE/RSJ International Conference on Intelligent Robots and Systems (IROS)}}, 2025.

\bibitem[Cao et~al.(2026)Cao, Chen, Maninis, Chen, Karpur, Xia, Dua, Dabral, Han, Han, et~al.]{cao2026tipsv2}
Bingyi Cao, Koert Chen, Kevis-Kokitsi Maninis, Kaifeng Chen, Arjun Karpur, Ye~Xia, Sahil Dua, Tanmaya Dabral, Guangxing Han, Bohyung Han, et~al.
\newblock {TIPSv2}: Advancing vision-language pretraining with enhanced patch-text alignment.
\newblock In \emph{{Proceedings of Conference on Computer Vision and Pattern Recognition (CVPR)}}, 2026.

\bibitem[Chen et~al.(2025)Chen, Guo, Zhu, Zhang, Huang, Feng, and Kang]{chen2025video}
Sili Chen, Hengkai Guo, Shengnan Zhu, Feihu Zhang, Zilong Huang, Jiashi Feng, and Bingyi Kang.
\newblock {Video Depth Anything}: Consistent depth estimation for super-long videos.
\newblock In \emph{{Proceedings of Conference on Computer Vision and Pattern Recognition (CVPR)}}, pp.\  22831--22840, 2025.

\bibitem[Chiang et~al.(2024)Chiang, Zheng, Sheng, Angelopoulos, Li, Li, Zhu, Zhang, Jordan, Gonzalez, and Stoica]{chiang2024chatbot}
Wei-Lin Chiang, Lianmin Zheng, Ying Sheng, Anastasios~Nikolas Angelopoulos, Tianle Li, Dacheng Li, Banghua Zhu, Hao Zhang, Michael Jordan, Joseph~E. Gonzalez, and Ion Stoica.
\newblock Chatbot {A}rena: An open platform for evaluating {LLMs} by human preference.
\newblock In \emph{{Proceedings of International Conference on Machine Learning (ICML)}}, pp.\  8359--8388, 2024.

\bibitem[Chow et~al.(2025)Chow, Mao, Li, Seita, Guizilini, and Wang]{chow2025physbench}
Wei Chow, Jiageng Mao, Boyi Li, Daniel Seita, Vitor~Campagnolo Guizilini, and Yue Wang.
\newblock {PhysBench}: Benchmarking and enhancing vision-language models for physical world understanding.
\newblock In \emph{{Proceedings of International Conference on Learning Representations (ICLR)}}, 2025.

\bibitem[Everingham et~al.(2010)Everingham, Van~Gool, Williams, Winn, and Zisserman]{everingham2010pascal}
Mark Everingham, Luc Van~Gool, Christopher K.~I. Williams, John Winn, and Andrew Zisserman.
\newblock The {PASCAL} visual object classes ({VOC}) challenge.
\newblock \emph{{International Journal of Computer Vision (IJCV)}}, 88\penalty0 (2):\penalty0 303--338, 2010.

\bibitem[Fang et~al.(2019)Fang, Li, Gao, and Jiang]{fang2019silly}
Yu~Fang, Minchen Li, Ming Gao, and Chenfanfu Jiang.
\newblock {Silly Rubber}: An implicit material point method for simulating non-equilibrated viscoelastic and elastoplastic solids.
\newblock \emph{ACM Transactions on Graphics (TOG)}, 38\penalty0 (4), 2019.
\newblock \doi{10.1145/3306346.3322968}.

\bibitem[Fang et~al.(2020)Fang, Qu, Li, Zhang, Zhu, Aanjaneya, and Jiang]{fang2020iqmpm}
Yu~Fang, Ziyin Qu, Minchen Li, Xinxin Zhang, Yixin Zhu, Mridul Aanjaneya, and Chenfanfu Jiang.
\newblock {IQ-MPM}: An interface quadrature material point method for non-sticky strongly two-way coupled nonlinear solids and fluids.
\newblock \emph{ACM Transactions on Graphics (TOG)}, 2020.

\bibitem[Feng et~al.(2026)Feng, Chen, Yu, Su, Terzopoulos, Yang, Masterjohn, Castro, and Jiang]{feng2026mpmlite}
Xiang Feng, Yunuo Chen, Chang Yu, Hao Su, Demetri Terzopoulos, Yin Yang, Joe Masterjohn, Alejandro Castro, and Chenfanfu Jiang.
\newblock {MPM Lite}: Linear kernels and integration without particles.
\newblock \emph{ACM Transactions on Graphics (TOG)}, 45\penalty0 (4), 2026.
\newblock \doi{10.1145/3811294}.

\bibitem[Gao et~al.(2026)Gao, Shu, Ye, Xiong, Makadia, Guo, Itti, and Chen]{gao20263dcodebench}
Yipeng Gao, Lei Shu, Genzhi Ye, Xi~Xiong, Ameesh Makadia, Meiqi Guo, Laurent Itti, and Jindong Chen.
\newblock {3DCodeBench}: Benchmarking agentic procedural {3D} modeling via code.
\newblock \emph{arXiv preprint arXiv:2606.01057}, 2026.

\bibitem[{Genesis AI Team}(2026)]{genesis2026genesisworld}
{Genesis AI Team}.
\newblock The role of simulation in scalable robotics, {Genesis World} 1.0, and the path forward.
\newblock \emph{Genesis AI Blog}, May 2026.
\newblock URL \url{https://www.genesis.ai/blog/the-role-of-simulation-in-scalable-robotics-genesis-world-10-and-the-path-forward}.

\bibitem[He et~al.(2026)He, Luo, Ma, and Averbuch-Elor]{he2026thinking}
Guangzhao He, Rundong Luo, Wei-Chiu Ma, and Hadar Averbuch-Elor.
\newblock Thinking in {Blender}: Staged executable inverse graphics with vision-language models.
\newblock \emph{arXiv preprint arXiv:2606.02580}, 2026.

\bibitem[Hu et~al.(2018)Hu, Fang, Ge, Qu, Zhu, Pradhana, and Jiang]{hu2018mlsmpm}
Yuanming Hu, Yu~Fang, Ziheng Ge, Ziyin Qu, Yixin Zhu, Andre Pradhana, and Chenfanfu Jiang.
\newblock A moving least squares material point method with displacement discontinuity and two-way rigid body coupling.
\newblock \emph{ACM Transactions on Graphics (TOG)}, 2018.

\bibitem[Hu et~al.(2019)Hu, Li, Anderson, Ragan-Kelley, and Durand]{hu2019taichi}
Yuanming Hu, Tzu-Mao Li, Luke Anderson, Jonathan Ragan-Kelley, and Fr{\'e}do Durand.
\newblock {Taichi}: A language for high-performance computation on spatially sparse data structures.
\newblock \emph{ACM Transactions on Graphics (TOG)}, 38\penalty0 (6), 2019.
\newblock \doi{10.1145/3355089.3356506}.

\bibitem[Hu et~al.(2020)Hu, Anderson, Li, Sun, Carr, Ragan-Kelley, and Durand]{hu2019difftaichi}
Yuanming Hu, Luke Anderson, Tzu-Mao Li, Qi~Sun, Nathan Carr, Jonathan Ragan-Kelley, and Fr{\'e}do Durand.
\newblock {DiffTaichi}: Differentiable programming for physical simulation.
\newblock In \emph{{Proceedings of International Conference on Learning Representations (ICLR)}}, 2020.

\bibitem[Hu et~al.(2021)Hu, Liu, Yang, Xu, Kuang, Xu, Dai, Freeman, and Durand]{hu2021quantaichi}
Yuanming Hu, Jiafeng Liu, Xuanda Yang, Mingkuan Xu, Ye~Kuang, Weiwei Xu, Qiang Dai, William~T. Freeman, and Fr{\'e}do Durand.
\newblock {QuanTaichi}: A compiler for quantized simulations.
\newblock \emph{ACM Transactions on Graphics (TOG)}, 40\penalty0 (4), 2021.
\newblock \doi{10.1145/3450626.3459671}.

\bibitem[Huang et~al.(2024)Huang, Chitalu, Lin, and Komura]{gipc2024}
Kemeng Huang, Floyd~M. Chitalu, Huancheng Lin, and Taku Komura.
\newblock {GIPC}: Fast and stable {G}auss-{N}ewton optimization of {IPC} barrier energy.
\newblock \emph{{ACM Transactions on Graphics (TOG)}}, 43\penalty0 (2), 2024.

\bibitem[Huang et~al.(2025)Huang, Lu, Lin, Komura, and Li]{huang2025stiffgipc}
Kemeng Huang, Xinyu Lu, Huancheng Lin, Taku Komura, and Minchen Li.
\newblock {StiffGIPC}: Advancing {GPU} {IPC} for stiff affine-deformable simulation.
\newblock \emph{ACM Transactions on Graphics (TOG)}, 44\penalty0 (3), 2025.
\newblock \doi{10.1145/3735126}.

\bibitem[Intern{\`o} et~al.(2026)Intern{\`o}, Pondaven, Issa, Pizzati, Pinto, Olhofer, Laptev, Torr, Simoncelli, Hammer, and Klindt]{interno2026geophys}
Christian Intern{\`o}, Alexander Pondaven, Habon Issa, Fabio Pizzati, Francesco Pinto, Markus Olhofer, Ivan Laptev, Philip Torr, Eero~P. Simoncelli, Barbara Hammer, and David Klindt.
\newblock {GEOPHYS}: The geometry of physical plausibility.
\newblock \emph{arXiv preprint arXiv:2606.20707}, 2026.

\bibitem[Kao et~al.(2026)Kao, Huynh, Wang, Vesdapunt, Stojanov, Hariharan, Obiednikov, and Zhou]{kao2026delta}
Chia-Hsiang Kao, Cong~Phuoc Huynh, Chien-Yi Wang, Noranart Vesdapunt, Stefan Stojanov, Bharath Hariharan, Oleksandr Obiednikov, and Ning Zhou.
\newblock {$\Delta$ynamics}: Language-based representation for inferring rigid-body dynamics from videos.
\newblock In \emph{{Proceedings of Conference on Computer Vision and Pattern Recognition (CVPR)}}, 2026.

\bibitem[Karaev et~al.(2025)Karaev, Makarov, Wang, Neverova, Vedaldi, and Rupprecht]{karaev2025cotracker3}
Nikita Karaev, Yuri Makarov, Jianyuan Wang, Natalia Neverova, Andrea Vedaldi, and Christian Rupprecht.
\newblock {CoTracker3}: Simpler and better point tracking by pseudo-labelling real videos.
\newblock In \emph{{Proceedings of International Conference on Computer Vision (ICCV)}}, 2025.

\bibitem[Kersten et~al.(2004)Kersten, Mamassian, and Yuille]{kersten2004object}
Daniel Kersten, Pascal Mamassian, and Alan Yuille.
\newblock Object perception as {B}ayesian inference.
\newblock \emph{Annual Review of Psychology}, 55:\penalty0 271--304, 2004.

\bibitem[Kl{\'a}r et~al.(2016)Kl{\'a}r, Gast, Pradhana, Fu, Schroeder, Jiang, and Teran]{klar2016sand}
Gergely Kl{\'a}r, Theodore Gast, Andre Pradhana, Chuyuan Fu, Craig Schroeder, Chenfanfu Jiang, and Joseph Teran.
\newblock {Drucker--Prager} elastoplasticity for sand animation.
\newblock \emph{ACM Transactions on Graphics (TOG)}, 35\penalty0 (4), 2016.
\newblock \doi{10.1145/2897824.2925906}.

\bibitem[Kong et~al.(2026)Kong, Li, Wang, Xu, Yao, Xiang, and Yang]{kong2026moge}
Lingyu Kong, Ruicheng Li, Ruicheng Wang, Sicheng Xu, Chengtang Yao, Jianfeng Xiang, and Jiaolong Yang.
\newblock {MoGe-3}: Fine-detail monocular geometry estimation with self-guided sparse volumetric refinement.
\newblock \emph{arXiv preprint arXiv:2607.17967}, 2026.

\bibitem[Kova{\v{c}}i{\v{c}} \& Ellis(2026)Kova{\v{c}}i{\v{c}} and Ellis]{kovavcivc2026mpmworlds}
{\v{Z}}iga Kova{\v{c}}i{\v{c}} and Kevin Ellis.
\newblock {MPMWorlds}: Material-point-method simulations for inferring and extrapolating physical dynamics.
\newblock \emph{arXiv preprint arXiv:2606.01538}, 2026.

\bibitem[Kulits et~al.(2024)Kulits, Feng, Liu, Fernandez~Abrevaya, and Black]{kulits2024rethinking}
Peter Kulits, Haiwen Feng, Weiyang Liu, Victoria Fernandez~Abrevaya, and Michael~J. Black.
\newblock Re-thinking inverse graphics with large language models.
\newblock \emph{Transactions on Machine Learning Research}, 2024.

\bibitem[Kulkarni et~al.(2015)Kulkarni, Kohli, Tenenbaum, and Mansinghka]{kulkarni2015picture}
Tejas~D. Kulkarni, Pushmeet Kohli, Joshua~B. Tenenbaum, and Vikash Mansinghka.
\newblock {Picture}: A probabilistic programming language for scene perception.
\newblock In \emph{{Proceedings of Conference on Computer Vision and Pattern Recognition (CVPR)}}, 2015.

\bibitem[Lan et~al.(2022)Lan, Kaufman, Li, Jiang, and Yang]{lan2022abd}
Lei Lan, Danny~M. Kaufman, Minchen Li, Chenfanfu Jiang, and Yin Yang.
\newblock Affine body dynamics: Fast, stable, and intersection-free simulation of stiff materials.
\newblock \emph{ACM Transactions on Graphics (TOG)}, 2022.

\bibitem[Li et~al.(2020)Li, Ferguson, Schneider, Langlois, Zorin, Panozzo, Jiang, and Kaufman]{li2020ipc}
Minchen Li, Zachary Ferguson, Teseo Schneider, Timothy Langlois, Denis Zorin, Daniele Panozzo, Chenfanfu Jiang, and Danny~M. Kaufman.
\newblock Incremental potential contact: Intersection- and inversion-free, large-deformation dynamics.
\newblock \emph{ACM Transactions on Graphics (TOG)}, 39\penalty0 (4), 2020.
\newblock \doi{10.1145/3386569.3392425}.

\bibitem[Li et~al.(2026)Li, Xiang, Mao, Wei, Chen, Masood, Li, and Adeli]{li2026quantiphy}
Puyin Li, Tiange Xiang, Ella Mao, Shirley Wei, Xinye Chen, Adnan Masood, Fei-Fei Li, and Ehsan Adeli.
\newblock {QuantiPhy}: A quantitative benchmark evaluating physical reasoning abilities of vision-language models.
\newblock In \emph{{Proceedings of Conference on Computer Vision and Pattern Recognition (CVPR)}}, 2026.

\bibitem[Li et~al.(2025)Li, Gu, Chen, Xu, Hu, Huang, and Wu]{li2025towards}
Wenqiao Li, Yao Gu, Xintao Chen, Xiaohao Xu, Ming Hu, Xiaonan Huang, and Yingna Wu.
\newblock Towards visual discrimination and reasoning of real-world physical dynamics: Physics-grounded anomaly detection.
\newblock In \emph{{Proceedings of Conference on Computer Vision and Pattern Recognition (CVPR)}}, 2025.

\bibitem[Liang et~al.(2026)Liang, Ku, Hui, Nie, and Chen]{liang2026visphyworld}
Jiarong Liang, Max Ku, Ka-Hei Hui, Ping Nie, and Wenhu Chen.
\newblock {VisPhyWorld}: Probing physical reasoning via code-driven video reconstruction.
\newblock \emph{arXiv preprint arXiv:2602.13294}, 2026.

\bibitem[Liang et~al.(2023)Liang, Bian, Xiao, Zhang, Chen, Liu, Xiang, Huang, and Su]{liang2023robo360}
Litian Liang, Liuyu Bian, Caiwei Xiao, Jialin Zhang, Linghao Chen, Isabella Liu, Fanbo Xiang, Zhiao Huang, and Hao Su.
\newblock {Robo360}: A {3D} omnispective multi-material robotic manipulation dataset.
\newblock \emph{arXiv preprint arXiv:2312.06686}, 2023.

\bibitem[Liu et~al.(2026)Liu, Wang, Chen, Zhao, Yang, and Jeni]{liu2026simworlds}
Chunjiang Liu, Xiaoyuan Wang, Haoyu Chen, Yizhou Zhao, Ming-Hsuan Yang, and L{\'a}szl{\'o}~A. Jeni.
\newblock {SimWorlds}: A multi-agent system for dynamic {3D} scene creation.
\newblock \emph{arXiv preprint arXiv:2607.01766}, 2026.

\bibitem[Liu et~al.(2025)Liu, Wang, and Li]{liu2025ckmpm}
Michael Liu, Xinlei Wang, and Minchen Li.
\newblock {CK-MPM}: A compact-kernel material point method.
\newblock \emph{ACM Transactions on Graphics (TOG)}, 44\penalty0 (4), 2025.
\newblock \doi{10.1145/3731155}.

\bibitem[Macklin(2022)]{Macklin_Warp_A_High-performance_2022}
Miles Macklin.
\newblock {Warp: A High-performance Python Framework for GPU Simulation and Graphics}, March 2022.
\newblock URL \url{https://github.com/NVIDIA/warp}.
\newblock NVIDIA GPU Technology Conference (GTC).

\bibitem[Motamed et~al.(2026)Motamed, Culp, Swersky, Jaini, and Geirhos]{motamed2026generative}
Saman Motamed, Laura Culp, Kevin Swersky, Priyank Jaini, and Robert Geirhos.
\newblock Do generative video models understand physical principles?
\newblock In \emph{{Proceedings of Winter Conference on Applications of Computer Vision (WACV)}}, 2026.

\bibitem[Niu et~al.(2026)Niu, Liu, Wang, Shao, Yin, Pai, Sharma, Saravalle, Zheng, Wang, et~al.]{niu2026t}
Dantong Niu, Zhuoyang Liu, Zekai Wang, Boning Shao, Zhao-Heng Yin, Anirudh Pai, Yuvan Sharma, Stefano Saravalle, Ruijie Zheng, Jing Wang, et~al.
\newblock {T-Rex}: Tactile-reactive dexterous manipulation.
\newblock \emph{arXiv preprint arXiv:2606.17055}, 2026.

\bibitem[Ranftl et~al.(2022)Ranftl, Lasinger, Hafner, Schindler, and Koltun]{ranftl2020towards}
Ren{\'e} Ranftl, Katrin Lasinger, David Hafner, Konrad Schindler, and Vladlen Koltun.
\newblock Towards robust monocular depth estimation: Mixing datasets for zero-shot cross-dataset transfer.
\newblock \emph{{Transactions on Pattern Analysis and Machine Intelligence (TPAMI)}}, 44\penalty0 (3):\penalty0 1623--1637, 2022.

\bibitem[Ritchie et~al.(2023)Ritchie, Guerrero, Jones, Mitra, Schulz, Willis, and Wu]{ritchie2023neurosymbolic}
Daniel Ritchie, Paul Guerrero, R.~Kenny Jones, Niloy~J. Mitra, Adriana Schulz, Karl D.~D. Willis, and Jiajun Wu.
\newblock Neurosymbolic models for computer graphics.
\newblock \emph{{Computer Graphics Forum}}, 42\penalty0 (2):\penalty0 545--568, 2023.

\bibitem[Shi et~al.(2023)Shi, Xu, Clarke, Li, and Wu]{shi2023robocook}
Haochen Shi, Huazhe Xu, Samuel Clarke, Yunzhu Li, and Jiajun Wu.
\newblock {RoboCook}: Long-horizon elasto-plastic object manipulation with diverse tools.
\newblock In \emph{{Conference on Robot Learning (CoRL)}}, volume 229, pp.\  642--660, 2023.

\bibitem[Shi et~al.(2024)Shi, Xu, Huang, Li, and Wu]{shi2024robocraft}
Haochen Shi, Huazhe Xu, Zhiao Huang, Yunzhu Li, and Jiajun Wu.
\newblock {RoboCraft}: Learning to see, simulate, and shape elasto-plastic objects in {3D} with graph networks.
\newblock \emph{The International Journal of Robotics Research}, 43\penalty0 (4):\penalty0 533--549, 2024.
\newblock \doi{10.1177/02783649231219020}.

\bibitem[Sim{\'e}oni et~al.(2025)Sim{\'e}oni, Vo, Seitzer, Baldassarre, Oquab, Jose, Khalidov, Szafraniec, Yi, Ramamonjisoa, et~al.]{simeoni2025dinov3}
Oriane Sim{\'e}oni, Huy~V. Vo, Maximilian Seitzer, Federico Baldassarre, Maxime Oquab, Cijo Jose, Vasil Khalidov, Marc Szafraniec, Seungeun Yi, Micha{\"e}l Ramamonjisoa, et~al.
\newblock {DINOv3}.
\newblock \emph{arXiv preprint arXiv:2508.10104}, 2025.

\bibitem[Spelke \& Kinzler(2007)Spelke and Kinzler]{spelke2007core}
Elizabeth~S. Spelke and Katherine~D. Kinzler.
\newblock Core knowledge.
\newblock \emph{Developmental Science}, 10\penalty0 (1):\penalty0 89--96, 2007.

\bibitem[Stomakhin et~al.(2013)Stomakhin, Schroeder, Chai, Teran, and Selle]{stomakhin2013snow}
Alexey Stomakhin, Craig Schroeder, Lawrence Chai, Joseph Teran, and Andrew Selle.
\newblock A material point method for snow simulation.
\newblock \emph{ACM Transactions on Graphics (TOG)}, 32\penalty0 (4), 2013.
\newblock \doi{10.1145/2461912.2461948}.

\bibitem[Sun et~al.(2026)Sun, Cao, Ping, Zhang, Chen, Xu, Ruan, Chen, Liu, Yan, et~al.]{sun2026maniparena}
Yu~Sun, Meng Cao, Yang Ping, Kaidong Zhang, Qingxuan Chen, Rongtao Xu, Liangwang Ruan, Xuecheng Chen, Dongxiu Liu, Yunxiao Yan, et~al.
\newblock {ManipArena}: Comprehensive real-world evaluation of reasoning-oriented generalist robot manipulation.
\newblock \emph{arXiv preprint arXiv:2603.28545}, 2026.

\bibitem[Suresh \& Atkeson(2026)Suresh and Atkeson]{suresh2026learning}
Krishna Suresh and Chris Atkeson.
\newblock Learning dynamic rope manipulation using task-level iterative learning control.
\newblock In \emph{{Robotics: Science and Systems (RSS)}}, 2026.

\bibitem[Tang et~al.(2026)Tang, Shimada, Meng, Bi, Liu, Wang, Xiao, Tan, Zhang, Huang, et~al.]{tang2026bvb}
Yolo~Y. Tang, Daiki Shimada, Jiayue Meng, Jing Bi, Pinxin Liu, Yicheng Wang, Yunzhong Xiao, Zhangyun Tan, Zeliang Zhang, Chao Huang, et~al.
\newblock {BVB}: Benchmarking agentic video understanding via programmatic reconstruction in {Blender}.
\newblock \emph{arXiv preprint arXiv:2609.15478}, 2026.

\bibitem[Teed \& Deng(2020)Teed and Deng]{teed2020raft}
Zachary Teed and Jia Deng.
\newblock {RAFT}: Recurrent all-pairs field transforms for optical flow.
\newblock In \emph{{Proceedings of European Conference on Computer Vision (ECCV)}}, 2020.

\bibitem[Tung et~al.(2023)Tung, Ding, Chen, Bear, Gan, Tenenbaum, Yamins, Fan, and Smith]{tung2023physion}
Hsiao-Yu Tung, Mingyu Ding, Zhenfang Chen, Daniel Bear, Chuang Gan, Josh Tenenbaum, Dan Yamins, Judith Fan, and Kevin Smith.
\newblock {Physion++}: Evaluating physical scene understanding that requires online inference of different physical properties.
\newblock In \emph{{Proceedings of the Neural Information Processing Systems (NeurIPS) Track on Datasets and Benchmarks}}, 2023.

\bibitem[Ullman et~al.(2017)Ullman, Spelke, Battaglia, and Tenenbaum]{ullman2017mind}
Tomer~D. Ullman, Elizabeth~S. Spelke, Peter Battaglia, and Joshua~B. Tenenbaum.
\newblock Mind games: Game engines as an architecture for intuitive physics.
\newblock \emph{Trends in Cognitive Sciences}, 21\penalty0 (9):\penalty0 649--665, 2017.

\bibitem[Umeyama(1991)]{umeyama1991least}
Shinji Umeyama.
\newblock Least-squares estimation of transformation parameters between two point patterns.
\newblock \emph{{Transactions on Pattern Analysis and Machine Intelligence (TPAMI)}}, 1991.

\bibitem[Wang et~al.(2026)Wang, Cai, Chen, Chi, Sun, Dai, Hung, Guo, Ren, Yao, et~al.]{wang2026code}
Hanyang Wang, Yimo Cai, Weiliang Chen, Jiawei Chi, Haowen Sun, Qiyu Dai, Yi-Hsin Hung, Xingzhuo Guo, Jinshan Ren, Runmao Yao, et~al.
\newblock {Code as Worlds}: Agentic discovery of executable world representations for physical reasoning.
\newblock \emph{arXiv preprint arXiv:2608.27549}, 2026.

\bibitem[Wang et~al.(2025)Wang, Ma, Cao, Zheng, Feng, Zhang, Pang, and Liang]{wang2025wisa}
Jing Wang, Ao~Ma, Ke~Cao, Jun Zheng, Jiasong Feng, Zhanjie Zhang, Wanyuan Pang, and Xiaodan Liang.
\newblock {WISA}: World simulator assistant for physics-aware text-to-video generation.
\newblock In \emph{{Proceedings of Advances in Neural Information Processing Systems (NeurIPS)}}, 2025.

\bibitem[Wang et~al.(2020)Wang, Li, Fang, Zhang, Gao, Tang, Kaufman, and Jiang]{wang2020hot}
Xinlei Wang, Minchen Li, Yu~Fang, Xinxin Zhang, Ming Gao, Min Tang, Danny~M. Kaufman, and Chenfanfu Jiang.
\newblock Hierarchical optimization time integration for {CFL}-rate {MPM} stepping.
\newblock \emph{ACM Transactions on Graphics (TOG)}, 39\penalty0 (3), 2020.
\newblock \doi{10.1145/3386760}.

\bibitem[Wolper et~al.(2019)Wolper, Fang, Li, Lu, Gao, and Jiang]{wolper2019cdmpm}
Joshuah Wolper, Yu~Fang, Minchen Li, Jiecong Lu, Ming Gao, and Chenfanfu Jiang.
\newblock {CD-MPM}: Continuum damage material point methods for dynamic fracture animation.
\newblock \emph{ACM Transactions on Graphics (TOG)}, 2019.

\bibitem[Wolper et~al.(2020)Wolper, Chen, Li, Fang, Qu, Lu, Cheng, and Jiang]{wolper2020anisompm}
Joshuah Wolper, Yunuo Chen, Minchen Li, Yu~Fang, Ziyin Qu, Jiecong Lu, Meggie Cheng, and Chenfanfu Jiang.
\newblock {AnisoMPM}: Animating anisotropic damage mechanics.
\newblock \emph{ACM Transactions on Graphics (TOG)}, 39\penalty0 (4), 2020.
\newblock \doi{10.1145/3386569.3392428}.

\bibitem[Wu et~al.(2015)Wu, Yildirim, Lim, Freeman, and Tenenbaum]{wu2015galileo}
Jiajun Wu, Ilker Yildirim, Joseph~J. Lim, Bill Freeman, and Josh Tenenbaum.
\newblock {Galileo}: Perceiving physical object properties by integrating a physics engine with deep learning.
\newblock In \emph{{Proceedings of Advances in Neural Information Processing Systems (NeurIPS)}}, 2015.

\bibitem[Wu et~al.(2016)Wu, Lim, Zhang, Tenenbaum, and Freeman]{wu2016physics}
Jiajun Wu, Joseph~J. Lim, Hongyi Zhang, Joshua~B. Tenenbaum, and William~T. Freeman.
\newblock {Physics 101}: Learning physical object properties from unlabeled videos.
\newblock In \emph{{British Machine Vision Conference (BMVC)}}, 2016.

\bibitem[Wu et~al.(2017{\natexlab{a}})Wu, Lu, Kohli, Freeman, and Tenenbaum]{wu2017learning}
Jiajun Wu, Erika Lu, Pushmeet Kohli, Bill Freeman, and Josh Tenenbaum.
\newblock Learning to see physics via visual de-animation.
\newblock In \emph{{Proceedings of Advances in Neural Information Processing Systems (NeurIPS)}}, 2017{\natexlab{a}}.

\bibitem[Wu et~al.(2017{\natexlab{b}})Wu, Tenenbaum, and Kohli]{wu2017neural}
Jiajun Wu, Joshua~B. Tenenbaum, and Pushmeet Kohli.
\newblock Neural scene de-rendering.
\newblock In \emph{{Proceedings of Conference on Computer Vision and Pattern Recognition (CVPR)}}, 2017{\natexlab{b}}.

\bibitem[Xie et~al.(2026)Xie, Wang, Hu, Qian, Wang, Wang, Ma, Peng, Wang, Tai, et~al.]{xie2026physcodebench}
Tianyidan Xie, Peiyu Wang, Jiaxin Hu, Yuyi Qian, Yuxuan Wang, Shenyi Wang, Rui Ma, Yanlun Peng, Lanjun Wang, Ying Tai, et~al.
\newblock {PhysCodeBench}: Benchmarking physics-aware symbolic simulation of {3D} scenes via self-corrective multi-agent refinement.
\newblock \emph{arXiv preprint arXiv:2604.23580}, 2026.

\bibitem[Yin et~al.(2026)Yin, Ge, Wang, Wang, Li, Black, Darrell, Kanazawa, and Feng]{yin2026vision}
Shaofeng Yin, Jiaxin Ge, Zora~Zhiruo Wang, Chenyang Wang, Xiuyu Li, Michael~J. Black, Trevor Darrell, Angjoo Kanazawa, and Haiwen Feng.
\newblock Vision-as-inverse-graphics agent via interleaved multimodal reasoning.
\newblock In \emph{{Proceedings of European Conference on Computer Vision (ECCV)}}, 2026.

\bibitem[Yuille \& Kersten(2006)Yuille and Kersten]{yuille2006vision}
Alan Yuille and Daniel Kersten.
\newblock Vision as {B}ayesian inference: analysis by synthesis?
\newblock \emph{Trends in Cognitive Sciences}, 10\penalty0 (7):\penalty0 301--308, 2006.

\bibitem[Zhang et~al.(2026)Zhang, Wang, Zhang, and Tang]{zhang2026code2worlds}
Yi~Zhang, Yunshuang Wang, Zeyu Zhang, and Hao Tang.
\newblock {Code2Worlds}: Empowering coding {LLMs} for {4D} world generation.
\newblock In \emph{{Proceedings of International Conference on Machine Learning (ICML)}}, 2026.

\bibitem[Zhao et~al.(2024)Zhao, Tompson, Driess, Florence, Ghasemipour, Finn, and Wahid]{zhao2024aloha}
Tony~Z. Zhao, Jonathan Tompson, Danny Driess, Pete Florence, Seyed Kamyar~Seyed Ghasemipour, Chelsea Finn, and Ayzaan Wahid.
\newblock {ALOHA} unleashed: A simple recipe for robot dexterity.
\newblock In \emph{{Conference on Robot Learning (CoRL)}}, 2024.

\bibitem[Zhao et~al.(2026)Zhao, Zhou, Yang, Tang, Jian, Yao, Yao, Lin, Guo, Chen, et~al.]{zhao2026sceneactbench}
Yifei Zhao, Xiangxin Zhou, Wenhao Yang, Jiaqi Tang, Pu~Jian, Huanjin Yao, Jiarui Yao, Haowei Lin, Chunchao Guo, Zhuo Chen, et~al.
\newblock {SceneActBench}: Can agents act on the {3D} scenes they see?
\newblock \emph{arXiv preprint arXiv:2607.22393}, 2026.

\bibitem[Zheng et~al.(2024)Zheng, Yin, Xie, Sun, Huang, Yu, Cao, Kozyrakis, Stoica, Gonzalez, Barrett, and Sheng]{zheng2024sglang}
Lianmin Zheng, Liangsheng Yin, Zhiqiang Xie, Chuyue Sun, Jeff Huang, Cody~Hao Yu, Shiyi Cao, Christos Kozyrakis, Ion Stoica, Joseph~E. Gonzalez, Clark Barrett, and Ying Sheng.
\newblock {SGLang}: Efficient execution of structured language model programs.
\newblock In \emph{{Proceedings of Advances in Neural Information Processing Systems (NeurIPS)}}, 2024.

\bibitem[Zhou et~al.(2024)Zhou, Wang, Ma, Liu, Huang, and Wang]{zhou2024uni3d}
Junsheng Zhou, Jinsheng Wang, Baorui Ma, Yu-Shen Liu, Tiejun Huang, and Xinlong Wang.
\newblock {Uni3D}: Exploring unified {3D} representation at scale.
\newblock In \emph{{Proceedings of International Conference on Learning Representations (ICLR)}}, 2024.

\end{thebibliography}
}
\newpage
\appendix
\setcounter{figure}{0}\setcounter{table}{0}
\renewcommand{\thefigure}{A\arabic{figure}}\renewcommand{\thetable}{A\arabic{table}}
\renewcommand{\theHfigure}{A\arabic{figure}}\renewcommand{\theHtable}{A\arabic{table}}
{%
    \centering
    {\scshape\fontsize{20}{25}\selectfont \textbf{\wordmark}: Benchmarking Agents on Inverse Graphics of Dynamic Scenes\par}
    \vspace{5mm}
    {\Large Appendix} \\[10mm]
}

\section*{\contentsname}
\startcontents[appendices]
\printcontents[appendices]{}{1}{\setcounter{tocdepth}{2}}

\section{Task Illustration}
\label{sec:supp:task}
\begin{figure}[ht]
    \centering
    \includegraphics[width=\textwidth]{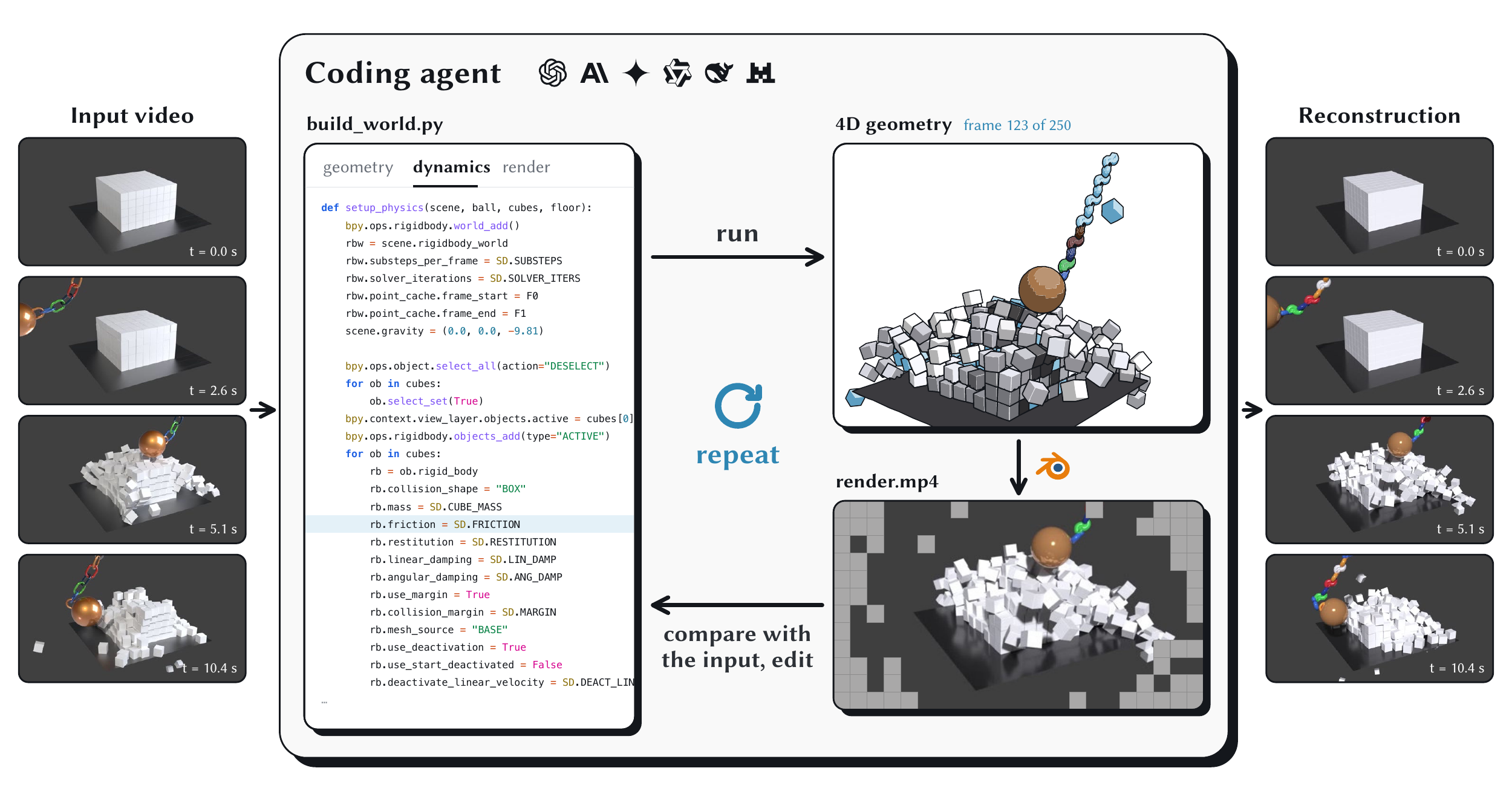}
    \caption{\textbf{Task.} Given the input video, the coding agent writes a program that builds a 4D scene, runs and renders it, compares the render with the input, and edits the program in a loop.}
    \label{fig:supp:task}
\end{figure}

\section{Additional Related Work}
\label{sec:supp:related_work}

\paragraph{Physical world understanding.} Humans reason about physical scenes by mentally simulating them with an approximate internal physics engine, grounded in core knowledge of objects and mechanics~\citep{spelke2007core,battaglia2013simulation,ullman2017mind}. Early computational frameworks such as Galileo and visual de-animation operationalize this principle by inferring structured, object-centric states from video and simulating them to predict and explain observed dynamics~\citep{wu2015galileo,wu2017learning}. Most contemporary benchmarks, however, probe physical reasoning through indirect surrogate tasks rather than explicit state representations, relying on outcome prediction and plausibility judgments~\citep{tung2023physion,bordes2025intphys}, question answering~\citep{chow2025physbench,li2026quantiphy}, or the physical plausibility of generated videos~\citep{bansal2025videophy,motamed2026generative}. Formulating scene understanding as world-program generation revives this explicit state representation, requiring agents to commit their physical understanding to an executable 4D scene that can be simulated, inspected, and verified.

\paragraph{Inverse graphics and executable world modeling.} Inverse graphics frames visual perception as inverting the forward rendering process through analysis-by-synthesis~\citep{kersten2004object,yuille2006vision}. Classical visual program induction pioneered this philosophy by recovering scenes as programs~\citep{kulkarni2015picture,wu2017neural,ritchie2023neurosymbolic}, but was largely constrained to hand-crafted domain-specific languages. Recent LLM-powered coding agents lift this limitation, parsing static objects and indoor layouts from imagery into general-purpose graphics code~\citep{kulits2024rethinking,yin2026vision,he2026thinking,gao20263dcodebench}. Extending such programs to dynamic scenes introduces a fundamentally more demanding inverse problem: the program must capture not only spatial geometry, but also how the physical scene evolves over time. Forward simulation from text~\citep{zhang2026code2worlds,liu2026simworlds,xie2026physcodebench} avoids this inverse challenge, as generated code needs only to satisfy an open-ended description rather than match a specific physical observation. Video-conditioned benchmarks directly tackle this inverse problem, yet each focuses on a restricted regime (Table~\ref{tab:supp:related_work}): VisPhyWorld and MPMWorlds operate on controlled synthetic simulations, with the latter confined to 2D~\citep{liang2026visphyworld,kovavcivc2026mpmworlds}; $\Delta$ynamics and SceneActBench target primarily rigid and articulated systems~\citep{kao2026delta,zhao2026sceneactbench}; and BVB reconstructs static indoor scenes under moving camera trajectories~\citep{tang2026bvb}. Concurrently, Code as Worlds builds executable scenes from real footage, but its video pipeline fits object trajectories kinematically rather than inferring the underlying physics~\citep{wang2026code}. By contrast, \BENCHMARK establishes a unified inverse-graphics benchmark across diverse real-world footage, spanning rigid, deformable, co-dimensional, and flowing matter, while evaluating the reconstructed 3D geometry and motion directly against ground truth on synthetic scenes.

\begin{table}[t]
\centering
\small
\setlength{\tabcolsep}{3pt}
\caption{Comparison with benchmarks for code-based world generation and reconstruction. Type: generation from text (Gen.) or reconstruction from images or video (Recon.). Real: real visual inputs. 3D geom.: the output contains 3D scene geometry. Dynamics columns mark which of our matter classes move in the tasks. 3D eval.: scores 3D geometry or simulator state rather than only rendered frames.}
\label{tab:supp:related_work}
\begin{tabular}{@{}llcccccccc@{}}
\toprule
& & & & \multicolumn{4}{c}{\textbf{Dynamics by matter class}} & \\
\cmidrule(lr){5-8}
\textbf{Benchmark} & \textbf{Type} & \textbf{Real} & \textbf{3D geom.} & \textbf{Rigid} & \textbf{Deform.} & \textbf{Co-dim.} & \textbf{Flowing} & \textbf{3D eval.} \\
\midrule
3DCodeBench~\citeyearpar{gao20263dcodebench} & Gen./Recon. & \xmark & \cmark & \xmark & \xmark & \xmark & \xmark & \cmark \\
PhysCodeBench~\citeyearpar{xie2026physcodebench} & Gen. & -- & \cmark & \cmark & \cmark & \cmark & \cmark & \cmark \\
Code4D~\citeyearpar{zhang2026code2worlds} & Gen. & -- & \cmark & \cmark & \cmark & \cmark & \cmark & \xmark \\
4DBuildBench~\citeyearpar{liu2026simworlds} & Gen. & -- & \cmark & \cmark & \cmark & \cmark & \cmark & \cmark \\
VisPhyBench~\citeyearpar{liang2026visphyworld} & Recon. & \xmark & \cmark & \cmark & \xmark & \xmark & \xmark & \cmark \\
MPMWorlds~\citeyearpar{kovavcivc2026mpmworlds} & Recon. & \xmark & \xmark & \cmark & \cmark & \xmark & \cmark & \xmark \\
$\Delta$ynamics~\citeyearpar{kao2026delta} & Recon. & \cmark & \cmark & \cmark & \xmark & \xmark & \xmark & \cmark \\
BVB~\citeyearpar{tang2026bvb} & Recon. & \cmark & \cmark & \xmark & \xmark & \xmark & \xmark & \xmark \\
SceneActBench~\citeyearpar{zhao2026sceneactbench} & Recon. & \xmark & \cmark & \cmark & \xmark & \xmark & \xmark & \cmark \\
\midrule
\textbf{Ours} & \textbf{Recon.} & \cmark & \cmark & \cmark & \cmark & \cmark & \cmark & \cmark \\
\bottomrule
\end{tabular}
\end{table}

\section{Dataset}
\label{sec:supp:dataset}
\begin{figure}[ht]
    \centering
    \includegraphics[width=\textwidth]{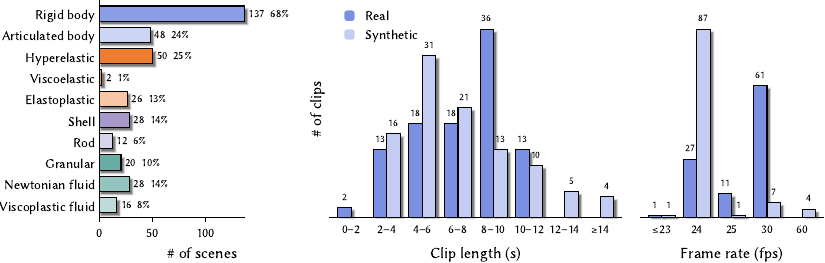}
    \caption{\textbf{Dataset distributions.} The number of scenes that contain each material and their proportion of the 200 scenes (left); since a scene can contain several materials, the shares sum to more than 100\%. The clip length (middle) and frame rate (right) distributions of the reference videos, shown separately for real and synthetic scenes.}
    \label{fig:supp:dataset_distributions}
\end{figure}

Figure~\ref{fig:supp:dataset_distributions} shows the number of scenes in \BENCHMARK{} that include an object of each material category (left) and the distributions of clip length and frame rate (middle, right).

\subsection{Dataset Samples}
\label{sec:supp:dataset_samples}
\begin{figure}[p]
    \centering
    \fullpagefigure{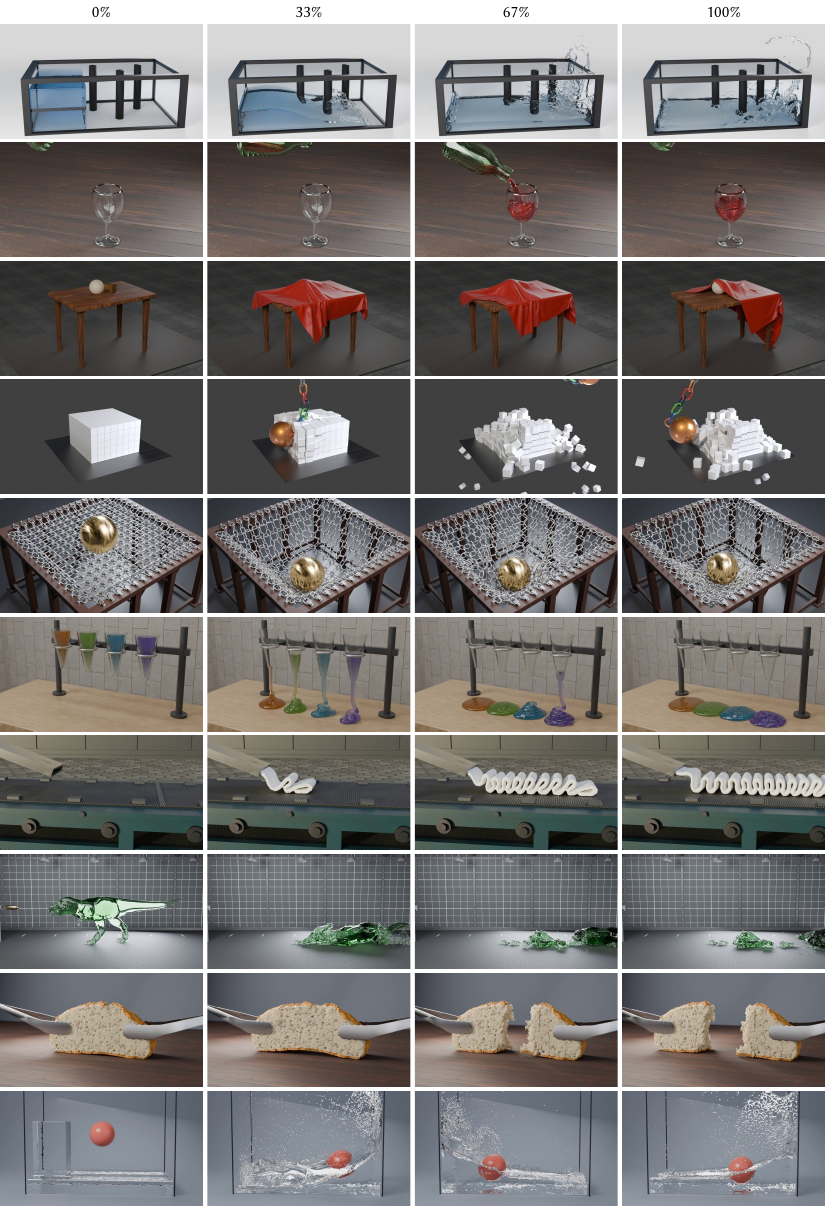}
    \caption{\textbf{Synthetic scenes (1/2).} Each scene is shown as four frames spaced evenly over its reference video.}
    \label{fig:supp:dataset_synthetic_1}
\end{figure}

\begin{figure}[p]
    \centering
    \fullpagefigure{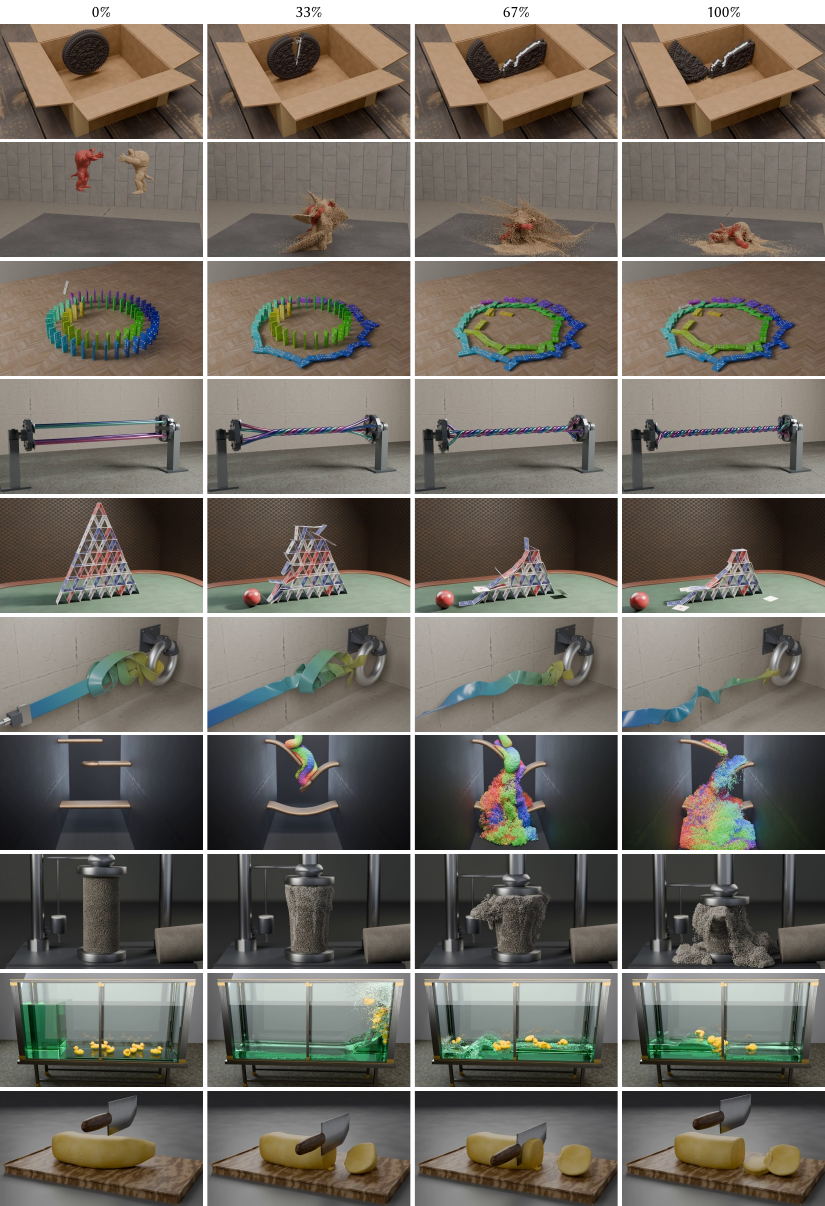}
    \caption{\textbf{Synthetic scenes (2/2).} Each scene is shown as four frames spaced evenly over its reference video.}
    \label{fig:supp:dataset_synthetic_2}
\end{figure}

\begin{figure}[p]
    \centering
    \fullpagefigure{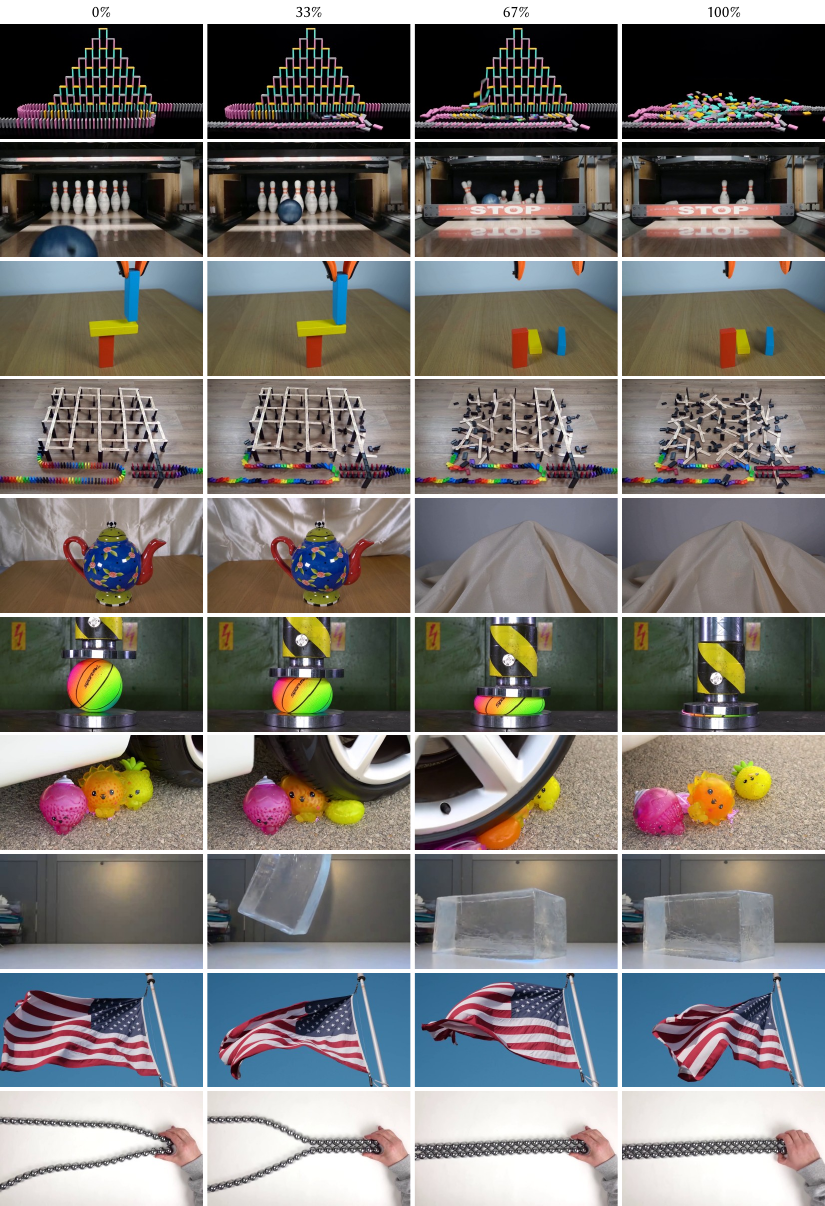}
    \caption{\textbf{Real-world scenes (1/2).} Each scene is shown as four frames spaced evenly over its reference video.}
    \label{fig:supp:dataset_real_1}
\end{figure}

\begin{figure}[p]
    \centering
    \fullpagefigure{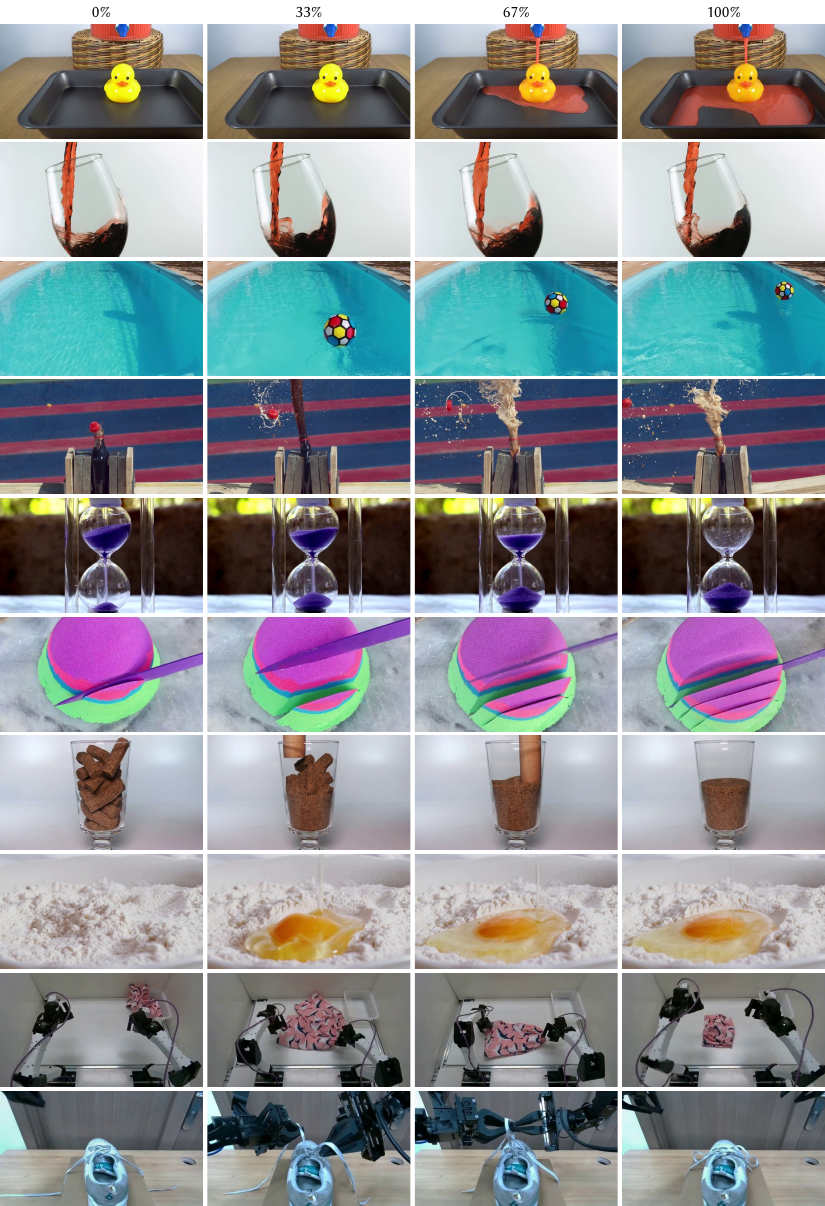}
    \caption{\textbf{Real-world scenes (2/2).} Each scene is shown as four frames spaced evenly over its reference video.}
    \label{fig:supp:dataset_real_2}
\end{figure}

Figures~\ref{fig:supp:dataset_synthetic_1}--\ref{fig:supp:dataset_real_2} show 40 of the 200 scenes: the synthetic pages cover every simulation method, and the real pages cover every physics category.

\subsection{Real-World Dataset Composition and Sources}
\label{sec:supp:real_data}

The real-world subset contains 100 video clips drawn from physics
video benchmarks (51 clips), robot-manipulation datasets (21 clips),
and independently collected web footage (28 clips). These sources
provide controlled physical demonstrations, robotic interactions
with objects and materials, and physical events in everyday settings.
Table~\ref{tab:supp:real_sources} lists the sources and their numbers.

\begin{table*}[ht]
\centering
\small
\setlength{\tabcolsep}{5pt}
\renewcommand{\arraystretch}{1.15}
\caption{\textbf{Real-world video sources} and the number of clips drawn from each.}
\label{tab:supp:real_sources}
\begin{tabular}{@{}llr@{}}
\toprule
\textbf{Source type} & \textbf{Source} & \textbf{Clips} \\
\midrule
Physics video benchmark
& WISA-80K~\citep{wang2025wisa} & 33 \\
& Physics-IQ~\citep{motamed2026generative} & 12 \\
& Phys-AD~\citep{li2025towards} & 3 \\
& Phys101~\citep{wu2016physics} & 3 \\
\addlinespace[5pt]
Robot manipulation
& ABC-130K~\citep{allshire2026scalable} & 5 \\
& T-Rex~\citep{niu2026t} & 4 \\
& Robo360~\citep{liang2023robo360} & 3 \\
& AgiBot World~\citep{bu2025agibot} & 2 \\
& RoboCook~\citep{shi2023robocook} & 2 \\
& RoboCraft~\citep{shi2024robocraft} & 2 \\
& ALOHA Unleashed~\citep{zhao2024aloha} & 1 \\
& ManipArena~\citep{sun2026maniparena} & 1 \\
& Task-Level ILC~\citep{suresh2026learning} & 1 \\
\addlinespace[5pt]
Web-collected 
& Pexels & 21 \\
& Youtube Creative Commons & 5 \\
& Mixkit & 1 \\
& Pixabay & 1 \\
\midrule
\textbf{Total} & & \textbf{100} \\
\bottomrule
\end{tabular}
\end{table*}

\subsection{Synthetic-Dataset Simulators}
\label{sec:supp:synthetic}
We use several simulation methods to cover the range of materials and interactions represented in the synthetic subset. At a high level, Incremental Potential Contact (IPC) formulates contact using barrier potentials to robustly prevent intersections during deformable simulation~\citep{li2020ipc}, while Affine Body Dynamics (ABD) extends related ideas to stiff and near-rigid bodies~\citep{lan2022abd}. Smoothed Particle Hydrodynamics (SPH) represents fluids using Lagrangian particles and is used for liquid dynamics, including fluid--rigid interactions and surface-tension effects. The Material Point Method (MPM) combines Lagrangian material points with an Eulerian background grid and supports a broad range of constitutive behaviors, including elasticity, plasticity, fracture, and coupled solid--fluid dynamics. We use several MPM variants specialized for different numerical and material regimes. Finally, the PPF Contact Solver uses finite-element models together with barrier-based contact handling for contact-rich simulations of shells, rods, and solids~\citep{ando2024cubic}. More than two-thirds of the synthetic examples are generated using a simulator that is more advanced and richer than Blender. 

 We extend the Genesis simulator~\citep{genesis2026genesisworld} with three new MPM constitutive models: a viscoplastic model to simulate materials such as shaving cream and toothpaste; the snow model of Stomakhin et al.~\citep{stomakhin2013snow}; and a Drucker--Prager model~\citep{klar2016sand} for sand. Sand and snow materials do exist in Genesis, but are implemented using different approaches than the ones we add.

\begin{table*}[ht]
\centering
\small
\setlength{\tabcolsep}{5pt}
\renewcommand{\arraystretch}{1.15}
\caption{\textbf{Simulators of the synthetic split}, the phenomena each one covers, and the number of clips it produced. All scenes are rendered in Blender.}
\label{tab:supp:simulators}
\begin{tabularx}{\textwidth}{@{}l
    >{\centering\arraybackslash}X
    r@{}}
\toprule
\textbf{Simulator} & \textbf{Simulation role} & \textbf{Clips} \\
\midrule
Blender
& Simpler physical scenes & 31 \\

libuipc~\citep{huang2025stiffgipc, gipc2024}
& Rigid, deformable, shell, and rod simulation with contact;
  affine-body assemblies & 10 \\
  
SPlisHSPlasH~\citep{SPlisHSPlasH_Library}
& Liquids, fluid--rigid coupling, and surface tension & 3 \\

\addlinespace[5pt]
MLS-MPM~\citep{hu2018mlsmpm}
& Cutting, discontinuities, and rigid-body coupling & 3 \\

CK-MPM~\citep{liu2025ckmpm}
& Deformable and elastoplastic materials & 9 \\

MPM-lite~\citep{feng2026mpmlite}
& Elastic and elastoplastic materials & 4 \\

HOT~\citep{wang2020hot}
& Stiff deformable and elastoplastic materials & 7 \\

\addlinespace[5pt]
AnisoMPM~\citep{wolper2020anisompm}
& Anisotropic damage and fracture; Drucker--Prager sand & 4 \\

CD-MPM~\citep{wolper2019cdmpm}
& Continuum-damage fracture & 4 \\

Silly Rubber~\citep{fang2019silly}
& Non-equilibrated viscoelastic and elastoplastic solids & 3 \\

IQ-MPM~\citep{fang2020iqmpm}
& Coupled deformable solids and fluids & 1 \\

\addlinespace[5pt]
PPF~\citep{ando2024cubic}
& Co-dimensional structures, deformables, and contact & 9 \\

Genesis~\citep{genesis2026genesisworld}
& Robotic manipulation, viscoplastic fluids, and MPM--SPH coupling & 12 \\
\midrule
\textbf{Total} & & \textbf{100} \\
\bottomrule
\end{tabularx}
\end{table*}

\section{Metrics and Evaluation}
\label{sec:supp:metrics}

\subsection{Executability and Scoring}
\label{sec:supp:executability}

\begin{figure}[t]
  \centering
  \includegraphics[width=0.65\linewidth]{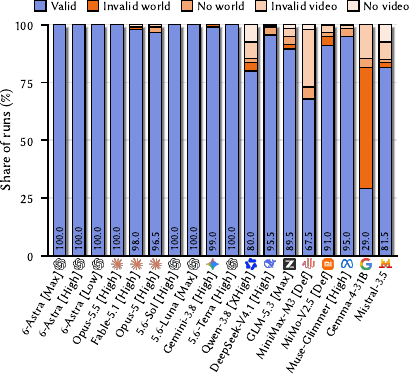}
  \caption{\textbf{Executability breakdown across models.}}
  \label{fig:executability}
\end{figure}

To ensure benchmark scores reflect both reconstruction fidelity and practical execution reliability, we first evaluate the validity of each submission.
A submission is designated as \emph{executable} if (i) its rendered video successfully decodes and matches the reference in resolution, frame count, and frame rate, and (ii) its 4D world passes structural format validation.
Submissions producing an invalid video automatically fail all metrics; those producing a valid video but an invalid 4D world fail only geometry-dependent metrics.
When a metric fails due to non-executability, it receives its worst possible value -- $0$ for bounded similarities and $2$ for normalized depth error (matching their bounds in Sections~\ref{sec:supp:metrics:all} and \ref{sec:supp:metrics:real}) -- so that aggregate scores penalize unexecutable outputs.
When the reference scene itself lacks the prerequisite signal for a specific metric (e.g., it contains no trackable dynamic motion), that scene is excluded from the corresponding metric across all models.
Surface rasterization fails to capture background details in six real scenes featuring transparent main objects. Consequently, these scenes are excluded from metrics comparing rasterized geometry to the reference video: Dynamic IoU, Depth error, Flow, Track2D, and Uni3D MoGe.
Figure~\ref{fig:executability} shows the executability of each model: 90.1\% of all submissions are executable, with at least 96.5\% for every proprietary model and 29.0--95.5\% for open-weight models.

\subsection{Metrics on All Scenes}
\label{sec:supp:metrics:all}
Let $x_k$ and $\hat{x}_k$ denote frame $k$ of the reference and rendered videos, respectively ($k=0,\ldots,T-1$).

\paragraph{DINOv3 similarity.}
To evaluate high-level visual appearance and semantic consistency across frames without being overly sensitive to minor pixel-level illumination or color shifts, we compute deep feature similarity using DINOv3~\citep{simeoni2025dinov3}. Using the class-token embeddings $e_k$ and $\hat{e}_k$ of the reference and rendered frames, we map the mean frame-wise cosine similarity to $[0,1]$:
\begin{equation}
\mathrm{DINOv3}=\frac{1}{2}\left(1+\frac{1}{T}\sum_{k}\frac{e_k^\top\hat{e}_k}{\|e_k\|\,\|\hat{e}_k\|}\right).
\end{equation}

\paragraph{Dynamic IoU.}
To capture the coarse spatial extent and general motion trends of dynamic elements across time, we measure Dynamic IoU. Let $\hat{M}_k$ denote the rasterized mask of dynamic objects defined by the submission's world program, and $M_k$ the reference dynamic mask, rasterized from the reference world or manually annotated for real videos. Following PASCAL VOC~\citep{everingham2010pascal}, we ignore a thin boundary band $B_k$ to avoid edge-rasterization artifacts:
\begin{equation}
\mathrm{DynIoU}=\frac{1}{T}\sum_{k}\frac{\left|(M_k\cap\hat{M}_k)\setminus B_k\right|}{\left|(M_k\cup\hat{M}_k)\setminus B_k\right|}.
\end{equation}

\subsection{Metrics on Real Videos}
\label{sec:supp:metrics:real}
For real videos, reference motion and depth are estimated from the source video, whereas reconstruction metrics are analytically rasterized directly from the predicted 4D world to assess true underlying geometry rather than rendered pictures.

\paragraph{Depth error.}
To evaluate 3D spatial layout and relative distance ordering from a monocular viewpoint, we compute normalized disparity error against estimated reference depth. Let $\tilde{d}_k$ and $\tilde{\hat{d}}_k$ be the disparities from Video Depth Anything~\citep{chen2025video} and the camera-rasterized reconstruction mesh, respectively, each median-and-MAD normalized as in MiDaS~\citep{ranftl2020towards}. Discarding the largest 1\% of residuals per frame to suppress outlier sensitivity, the pixel-averaged error over $N$ pixels is:
\begin{equation}
\mathrm{DepthErr}=\frac{1}{T}\sum_{k}\frac{1}{N}\sum_{p}\left|\tilde{d}_k(p)-\tilde{\hat{d}}_k(p)\right|,
\end{equation}
which is bounded in $[0,2]$.

\paragraph{Flow.}
To assess 2D apparent motion dynamics and velocity distributions across consecutive frames, we compare optical flow fields. Let $U_k$ be the reference RAFT~\citep{teed2020raft} flow, and $\hat{U}_k$ the reconstruction flow derived analytically from surface geometry, between frames $k$ and $k+1$ at the pixels that move in either, and $\mathcal{K}$ the steps at which any pixel moves. Rather than penalizing minor spatial misalignments with rigid pixel-wise end-point error, we compare them as unordered sets at each motion step $k \in \mathcal{K}$ using the sliced Wasserstein distance ($\mathrm{SW}$), a fast approximation of the earth mover's distance (EMD). Normalizing by the root-mean-square magnitude $m$ of each side ensures slow and fast scenes share one common scale:
\begin{equation}
\mathrm{Flow}=1-\frac{1}{|\mathcal{K}|}\sum_{k\in\mathcal{K}}\min\left(1,\frac{\mathrm{SW}(U_k,\hat{U}_k)}{m(U_k)+m(\hat{U}_k)}\right).
\end{equation}

\paragraph{Track2D.}
To evaluate long-range temporal motion persistence and tracking fidelity over extended sequences, we compute 2D point trajectory concordance. CoTracker3~\citep{karaev2025cotracker3} tracks a grid of query points through the reference video. On the reconstruction, the corresponding 3D surface point under each query is followed through the 4D world and projected into the frame, so path $i$ on one side corresponds to path $i$ on the other. Retaining the $n$ reference paths that travel more than $2\%$ of the image diagonal $\delta$ after camera motion compensation, we compute the Dynamic Time Warping (DTW) distance $D_i$ between corresponding paths in units of $\delta$, which accommodates variations in execution speed, capped at $\kappa=0.03$:
\begin{equation}
\mathrm{Track2D}=1-\frac{1}{n}\sum_{i}\frac{\min(D_i,\kappa)}{\kappa}.
\end{equation}

\subsection{Metrics on Synthetic Scenes}
\label{sec:supp:metrics:synthetic}
Because synthetic scenes provide complete ground-truth geometry, the metrics below are evaluated directly in 3D after registering the reconstruction to the reference world.

\paragraph{Registration.}
Let $\mathcal{P}$ be the point cloud of surfaces visible at frame $0$ of the reference world, with RMS radius $\sigma$; all 3D distances are measured in units of $\sigma$. Because lower-performing models frequently produce erroneous camera extrinsics due to subtle Blender camera export API conventions, relying solely on camera-view surfaces can lead to catastrophic misalignment despite otherwise plausible reconstruction. To prevent these models from failing 3D metrics purely due to misplaced cameras, we fit two similarity transforms by trimmed ICP~\citep{besl1992method,umeyama1991least} -- one on the visible surfaces at frame $0$ and one on the dynamic objects' trajectories across the video -- and keep the one that brings the reconstruction closer to $\mathcal{P}$ through the reference camera.

\paragraph{Scene 3D.}
To measure static 3D geometric fidelity and surface reconstruction accuracy under full 3D supervision, we evaluate the aligned frame-$0$ reconstruction against the reference ground truth. Let $\hat{\mathcal{P}}$ be the registered reconstruction's frame-$0$ surfaces seen through the reference camera, and $D_\kappa$ the symmetric Chamfer distance capped at $\kappa=0.5$:
\begin{equation}
\mathrm{Scene3D}=1-D_\kappa(\mathcal{P},\hat{\mathcal{P}})/\kappa.
\end{equation}

\paragraph{Trajectory DTW.}
To evaluate Lagrangian motion consistency -- measuring whether distinct physical matter follows the correct 3D trajectory over time -- we compute path-level dynamic time warping. We sample dynamic matter across both worlds uniformly and follow each sample over time, giving two sets of 3D paths $\{\tau_i\}$ and $\{\hat{\tau}_j\}$ of equal size $n$. We solve for the optimal one-to-one assignment $\pi$ with the Hungarian algorithm and, with saturation threshold $\kappa=1$, compute:
\begin{equation}
\mathrm{TrajDTW}=1-\frac{1}{n}\sum_{i}\frac{\min\big(\mathrm{DTW}(\tau_i,\hat{\tau}_{\pi(i)}),\kappa\big)}{\kappa}.
\end{equation}

\paragraph{EMD step.}
To evaluate Eulerian instantaneous velocity fields---capturing overall 3D physical speed and directional plausibility at each frame without requiring persistent particle tracking---we measure one-step displacement distributions. Let $U_k$ and $\hat{U}_k$ be the one-step 3D displacements of the same samples at frame $k$. To evaluate instantaneous velocity distributions without enforcing trajectory-level temporal persistence, we measure their sliced Wasserstein distance ($\mathrm{SW}$):
\begin{equation}
\mathrm{EMD\ step}=\max\left(0,\ 1-\frac{\sum_k\mathrm{SW}(U_k,\hat{U}_k)}{\sum_k\big(m(U_k)+m(\hat{U}_k)\big)}\right).
\end{equation}
Normalizing by the total motion over the video prevents nearly static frames from dominating the score.

\subsection{VQA}
\label{sec:supp:vqa}
Each scene carries hand-written binary questions in four categories: \emph{initial
state} and \emph{final state} (what is present and where, in the first and last
frames), \emph{key events} (what happens in between, and in what order), and
\emph{contact} (what the materials do where they touch). Over the 200 scenes there are
1,251 questions: 371 initial state, 354 final
state, 308 key events, and 218 contact.
A VLM judge (\texttt{Gemini 3.8 Flash}) sees 16 frames of the
reconstruction -- including the first and last frames -- and answers all of a scene's questions in one call, without the reference video.

\paragraph{Gating.} Two checks run before the VLM is asked to answer the questions. The first catches renders with nothing in them: a render fails if it has almost no edges, or if almost every pixel is
the same color. These are blank, black, or flat-lit outputs. The second check is per scene. Every scene carries a presence question about its main object, and if the reconstruction fails it, the object is not in the video and the remaining answers say nothing about physics. The question essentially asks if the reconstruction contains anything that could even remotely try to represent the main dynamic object of the scene.

\begin{figure*}[p]
    \centering
    \halfpagefigure{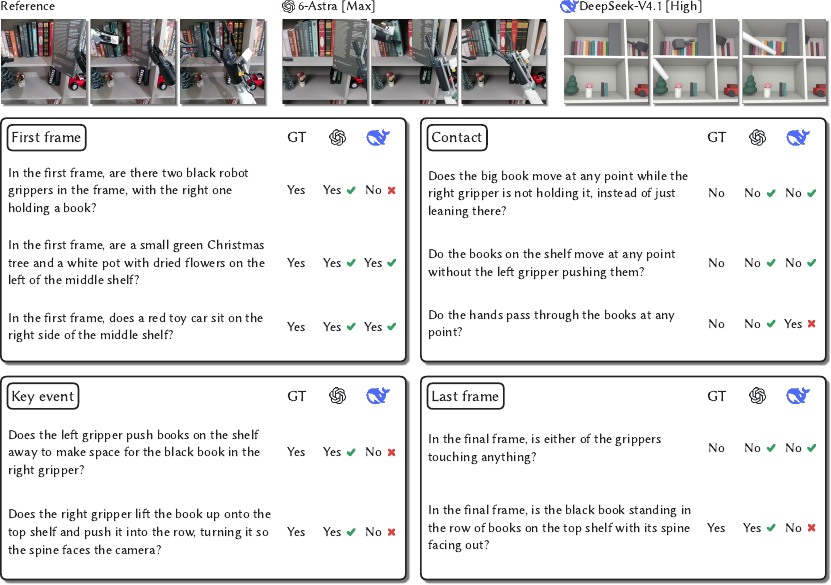}

    \vspace{1em}
    \halfpagefigure{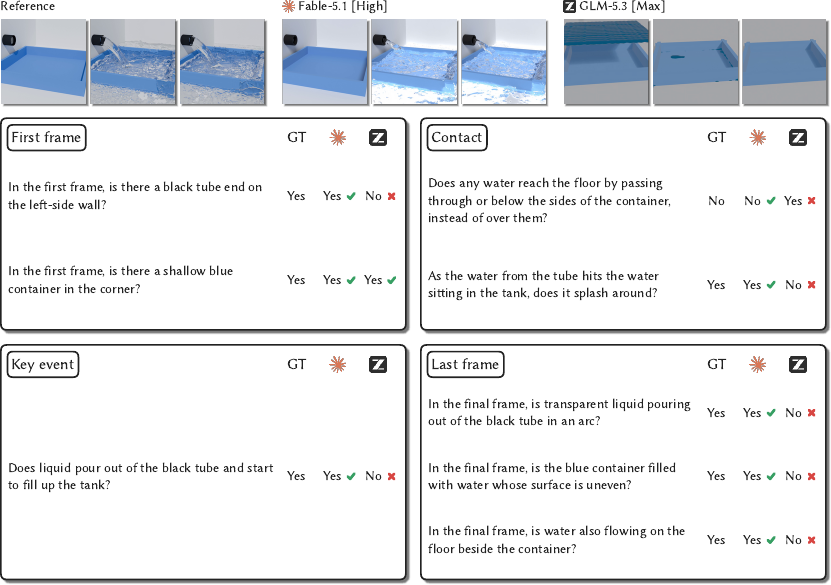}
    \caption{\textbf{VQA examples on a real rigid (top) and a synthetic flowing (bottom) scene.} Each shows frames at the start, middle, and end of the reference video and of two reconstructions, and up to three questions per category with the ground-truth (GT) answer and each model's answer, marked right or wrong.}
    \label{fig:supp:vqa_examples}
\end{figure*}

We score the questions of a gated scene as wrong rather than skipping them, and we do the
same for scenes that produced no video. Every model therefore answers the same 1,251 questions, so no model can raise its average by rendering fewer scenes. Across all models, 81.0\% of scenes reach
the judge cleanly, 3.6\% produce no video, 8.5\% fail the first check, and 7.0\% fail the presence question.
Figure~\ref{fig:supp:vqa_examples} shows the questions of two scenes with the answers of two models on each.

\subsection{VLM-Based Pairwise Rankings} 
\label{sec:vlm_pairwise}
For each comparison, the judge receives the reference video and two model reconstructions, sampled at 3\,fps. The model used is \texttt{Gemini 3.8 Flash}. The VLM chooses which reconstruction more closely matches the reference in object geometry, motion, collisions, deformation, and final state. Model identities are hidden and A/B assignment is balanced across comparisons.

We sample model pairs across scenes rather than exhaustively evaluating every pair. Pairwise outcomes are aggregated with a Bradley--Terry model and reported on the Elo scale. Confidence intervals are obtained by bootstrapping scenes.

\begin{figure}[ht]
    \centering
     \includegraphics[width=0.5\linewidth]{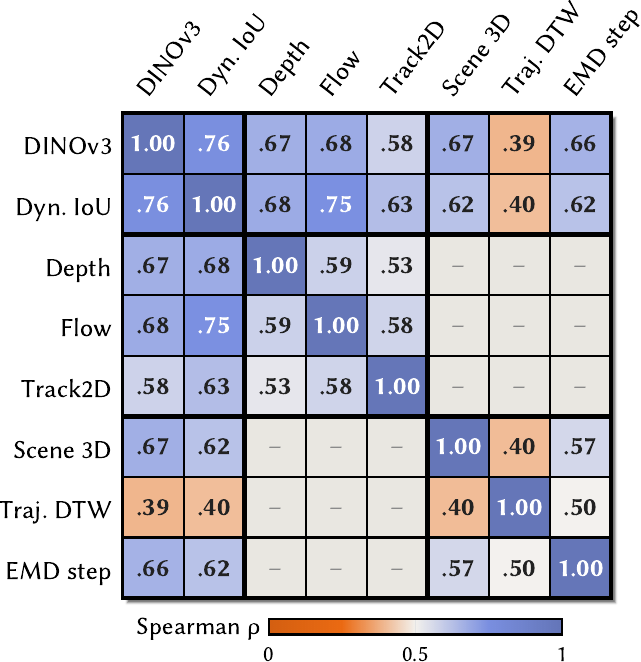}
    \caption{\textbf{Metric concordance.} Each cell is Spearman's $\rho$ between the rankings that two metrics assign to the runs on a scene, averaged over scenes; metrics are oriented so that higher is better, and grey pairs share no scene.}
    \label{fig:supp:metric_concordance}
\end{figure}

\begin{tcolorbox}[enhanced, breakable, colback=gray!4, colframe=gray!55,
    boxrule=0.5pt, arc=2pt, left=5pt, right=5pt, top=4pt, bottom=4pt,
    title={Pairwise judge prompt}, coltitle=black,
    colbacktitle=gray!15, fonttitle=\footnotesize\bfseries]
\begin{Verbatim}[fontsize=\scriptsize, breaklines]
You are shown a reference recording of a physical event, then two
attempts by coding models to reproduce it: video A and video B.

Which attempt is closer to the reference? Consider the geometry of
the objects in the scene, how they move, the physics, the
collisions, the deformations, where things end up.

Answer with one word: A or B.
\end{Verbatim}
\end{tcolorbox}

\subsection{Metric Concordance}

Figure~\ref{fig:supp:metric_concordance} evaluates whether two metrics rank submissions to the same scene consistently. Every pair agrees positively, with $\rho$ between 0.39 and 0.76, indicating that the metrics point in a common direction without being redundant. Appearance and layout agree closely: DINOv3 similarity and Dynamic IoU correlate at 0.76, and both at 0.67--0.68 with depth error. The motion metrics agree only moderately with each other and with appearance (0.53--0.69), indicating that a rendered video can look plausible while its underlying motion is inaccurate. Trajectory DTW is the most independent metric (0.39--0.50), since it uniquely requires the correct matter to follow the correct 3D path, which neither appearance nor per-frame motion statistics reveal.

\subsection{Additional Metrics and Geometry Quality}
\label{sec:supp:additional_metrics}
\begin{figure*}[ht]
    \centering
    \includegraphics[width=\textwidth]{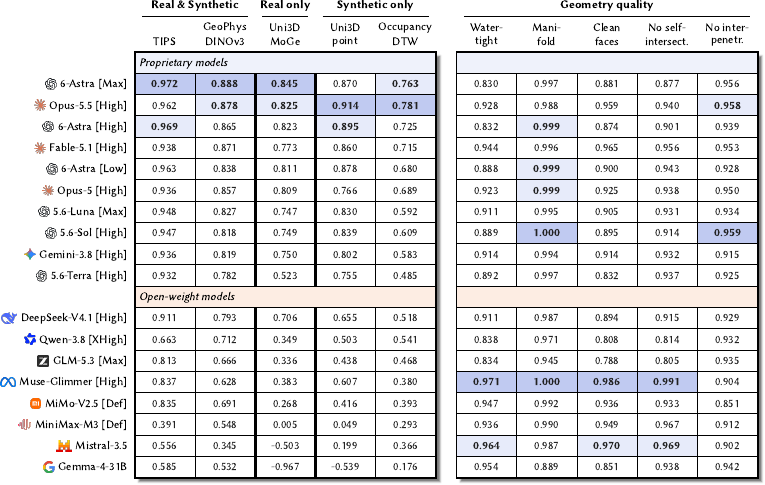}
\caption{\textbf{Additional metrics and geometry quality.} The left table scores five further reconstruction metrics, counting failed runs at the penalty value, and the right table scores the geometry checks over the runs that deliver a world; higher is better, and the best and second best in each column are shaded. The geometry checks do not follow reconstruction quality: weaker models, which tend to build simple geometry, often score higher on them.}
\label{fig:supp:additional_metrics}
\end{figure*}

Figure~\ref{fig:supp:additional_metrics} (left) reports five complementary metrics that broadly corroborate the primary benchmark rankings:
(i) \emph{TIPS similarity} is the same semantic similarity, computed with TIPSv2~\citep{cao2026tipsv2} embeddings in place of DINOv3;
(ii) \emph{GeoPhys DINOv3} evaluates temporal smoothness using five trajectory regularity descriptors from GeoPhys~\citep{interno2026geophys}, scoring $1-\frac{1}{5}\sum_{\phi}|\phi-\hat{\phi}|/(\phi+\hat{\phi})$;
(iii) \emph{Uni3D point} and \emph{Uni3D MoGe} compute cosine similarities of Uni3D~\citep{zhou2024uni3d} embeddings between reconstruction frame-$0$ surfaces and, respectively, the registered reference world and a single-frame MoGe~\citep{kong2026moge} point map;
and (iv) \emph{Occupancy DTW} evaluates dynamic matter sequence matching under sliced Wasserstein distance without requiring persistent particle identity.

\needspace{14\baselineskip}
\begin{wrapfigure}{r}{0.54\textwidth}
    \vspace{-1em}
    \centering
    \includegraphics[width=\linewidth]{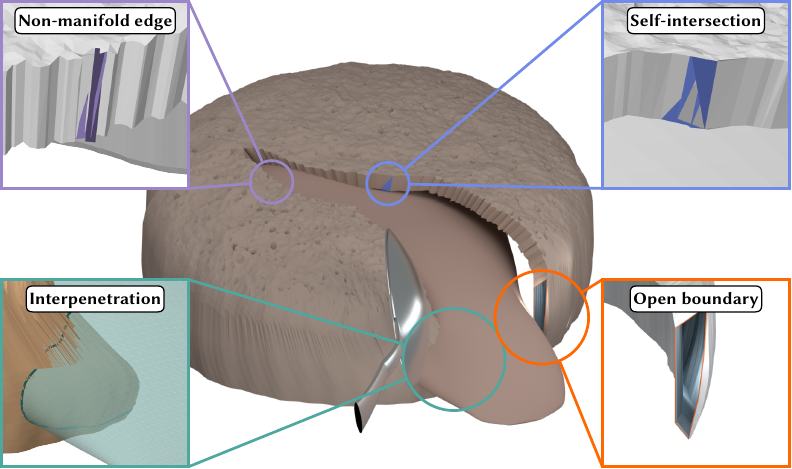}
    \caption{\textbf{Examples of geometry defects} on a reconstruction of a spoon splitting a lava cake at frame 150; each circle is magnified in the inset of the same color.}
    \label{fig:supp:geometry_defects}
    \vspace{-1em}
\end{wrapfigure}
\paragraph{Geometry quality.}
Beyond visual and dynamic tracking, we evaluate the physical validity of the reconstructed meshes.
Figure~\ref{fig:supp:geometry_defects} illustrates common structural defects (e.g., self-intersections and non-manifold edges).
In Figure~\ref{fig:supp:additional_metrics} (right), we report the fraction of mesh components across all frames that are \emph{watertight}, \emph{manifold}, \emph{non-degenerate}, and \emph{free of self-intersections}, alongside \emph{no interpenetration} ($1 -$ vertex penetration ratio).
These structural checks evaluate only submissions that successfully deliver a valid 4D world (executability rates in Figure~\ref{fig:executability}).

Notably, mesh quality does not correlate positively with reconstruction fidelity.
For instance, Muse Glimmer [High] leads or ties on all four mesh integrity checks, whereas Astra [Max]---the top-performing reconstruction model---achieves only 0.830 watertightness.
This divergence arises because stronger models attempt to reconstruct highly intricate geometry (such as fractured fragments, thin shells, and fluid surfaces) that inherently poses severe meshing challenges, whereas weaker models generate simple geometric primitives that pass structural tests effortlessly.
Interpenetration remains uniformly low across all models (0.851--0.959).

\section{Human Study}
\label{sec:supp:human_study}

\begin{figure*}[t]
    \centering
    \begin{minipage}[c]{0.46\textwidth}
    \begin{tcolorbox}[enhanced, colback=gray!4, colframe=gray!55, boxrule=0.5pt, arc=2pt, left=5pt, right=5pt, top=4pt, bottom=4pt,
        title={Which video looks better?}, coltitle=black, colbacktitle=gray!15, fonttitle=\footnotesize\bfseries, fontupper=\scriptsize]
    Each trial shows a \emph{reference} clip and two reconstructions of the same scene.

    \smallskip
    Your task is to pick the reconstruction that matches the reference more closely in \textbf{geometry} (shapes and sizes), \textbf{layout} (object positions, orientations, and viewpoint), and \textbf{motion} (movement paths, speed, and the order of events). Also consider whether the objects behave in a \textbf{physically plausible} way.

    \smallskip
    Watch the reference first; and then the two reconstructions will appear below and play together. After that you could drag the bar to view freely. The answer is one of \textbf{Left is better}, \textbf{About the same} and \textbf{Right is better}.

    \smallskip
    One practice trial, then 50 trials. The expected completion time is 21 minutes.
    \end{tcolorbox}
    \end{minipage}\hfill
    \begin{minipage}[c]{0.50\textwidth}
        \centering
        \includegraphics[width=\linewidth]{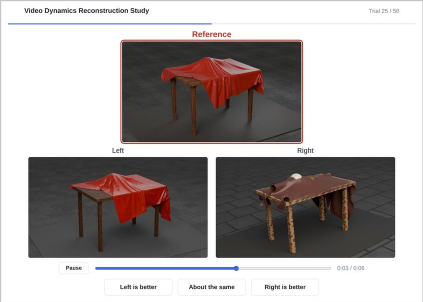}
    \end{minipage}
    \caption{\textbf{Human evaluation interface.} \emph{Left:} Instructions provided to participants outlining criteria for geometry, layout, motion, and physical plausibility. \emph{Right:} The side-by-side trial interface showing the reference video above two candidate reconstructions, equipped with synchronized scrub bars and ternary voting buttons (\emph{left}, \emph{right}, \emph{about the same}).}
    \label{fig:supp:study_interface}
\end{figure*}
\paragraph{Protocol.}
We conduct a side-by-side user study to measure which reconstruction people judge closer to the reference.
In each trial, participants view the reference video alongside rendered outputs from two randomly ordered models and, after replaying and scrubbing, select the one that matches the reference more closely in geometry, layout, and motion (\emph{left}, \emph{right}, or \emph{about the same}).
A session contains 49 comparisons on distinct scenes, balanced between real and synthetic splits, plus a practice trial and one comparison repeated with the sides swapped to evaluate self-consistency; participants give the same answer on 95.5\% of the 67 repeated comparisons. Figure~\ref{fig:supp:study_interface} shows the instructions and the interface.

\paragraph{Sampling.}
With 17 models, uniform sampling would waste judgments on pairs whose relative ranking is already certain. To concentrate the budget where rankings remain ambiguous, each trial selects the least-evaluated scene and samples model pair $(a,b)$ from its candidate pool by expected confidence interval shrinkage~\citep{chiang2024chatbot}:
\begin{equation}
p_{ab}\propto\sqrt{\hat{v}_{ab}/n_{ab}}-\sqrt{\hat{v}_{ab}/(n_{ab}+1)},
\end{equation}
where $n_{ab}$ and $\hat{v}_{ab}$ are the number and variance of the pair's past outcomes.

\paragraph{Ratings.}
We fit a Bradley--Terry model~\citep{bradley1952rank} on the Elo scale:
\begin{equation}
P(a\succ b)=\frac{1}{1+10^{-(R_a-R_b)/400}},
\end{equation}
via weighted maximum likelihood, applying importance weights $1/(K\,p_{ab})$ to correct for non-uniform sampling ($K$ denoting candidate pairs per scene) and counting ties as half-wins. Confidence intervals are obtained via 100 bootstrap resamples. Human Elo ratings correlate strongly with our aggregate automated ranking, the Overall score of Figure~\ref{fig:main_leaderboard} (Spearman's $\rho=0.96$).

\clearpage
\section{Additional Results}
\label{sec:supp:additional_results}
\subsection{Sample Agent Results}
\begin{center}
    \captionsetup{type=figure}
    \includegraphics[width=\textwidth,height=0.84\textheight,keepaspectratio]{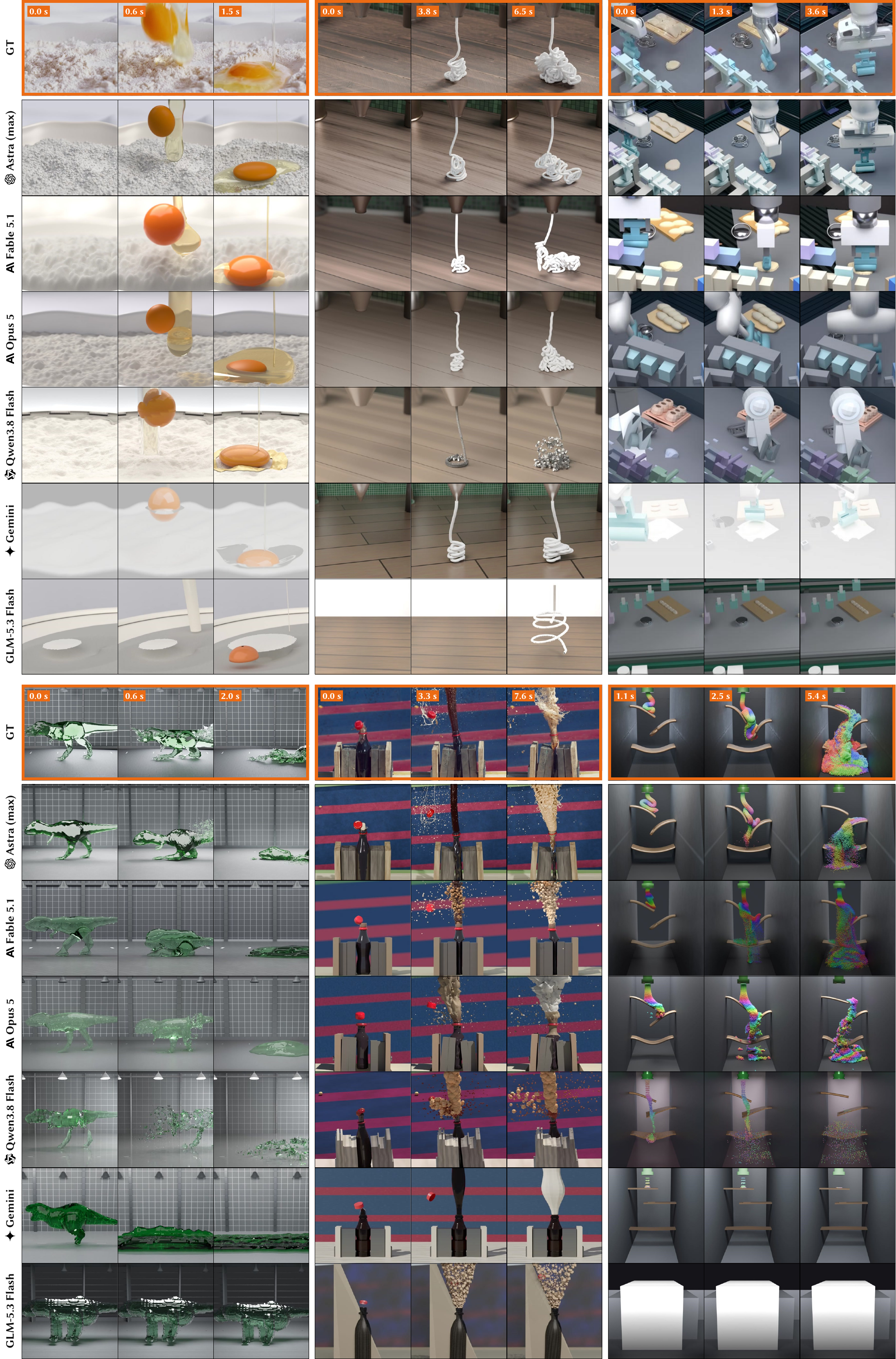}
    \caption{\textbf{Model Results.} This figure shows additional results from 6 models on 6 different \BENCHMARK tasks.}
    \label{fig:supp:results_supp_1}
\end{center}

\begin{figure}[p]
    \centering
    \fullpagefigure{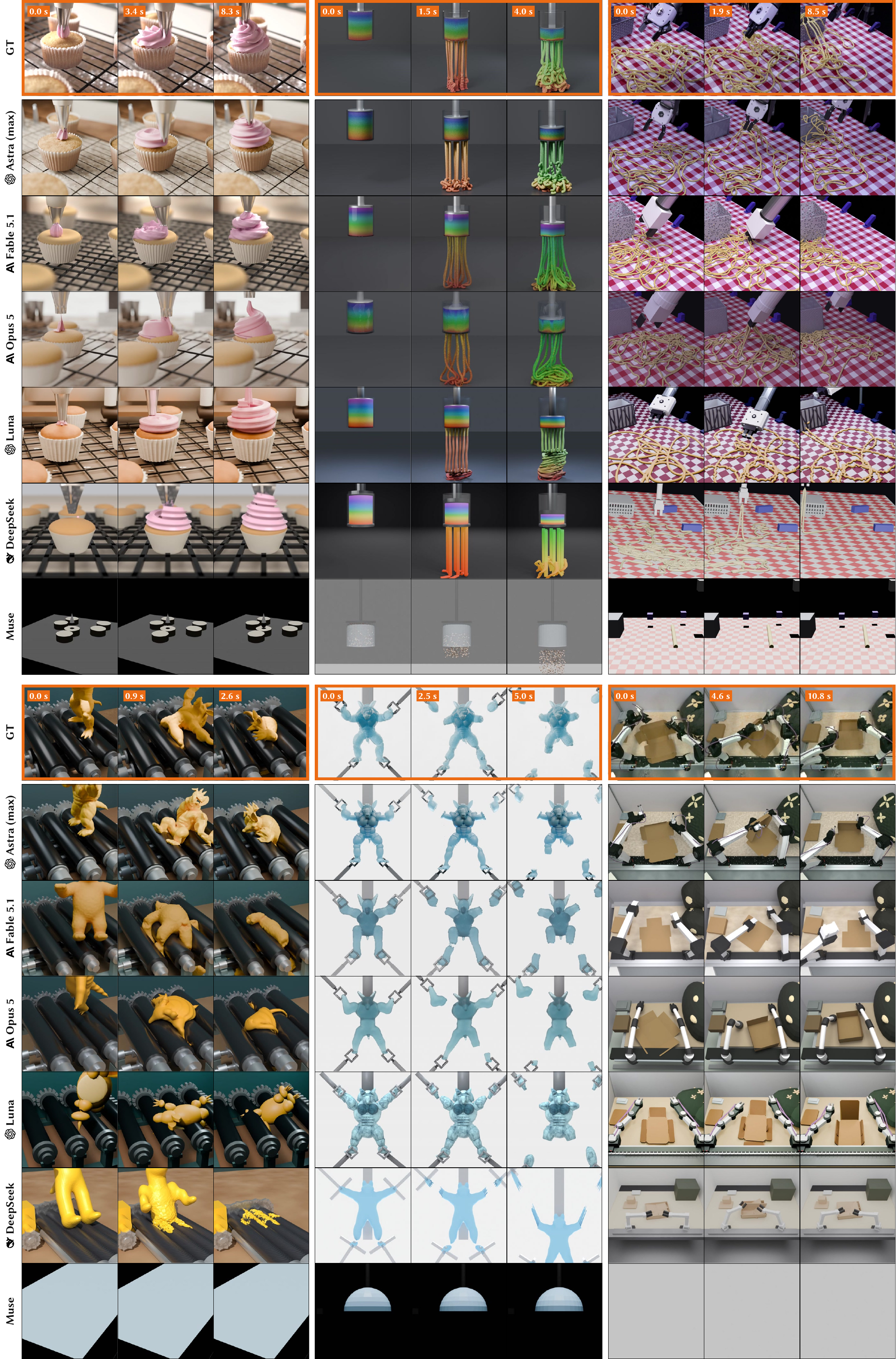}
    \caption{\textbf{Model Results.} This figure shows additional results from 6 models on 6 different \BENCHMARK tasks.}
    \label{fig:supp:results_supp_2}
\end{figure}

\begin{figure}[p]
    \centering
    \fullpagefigure{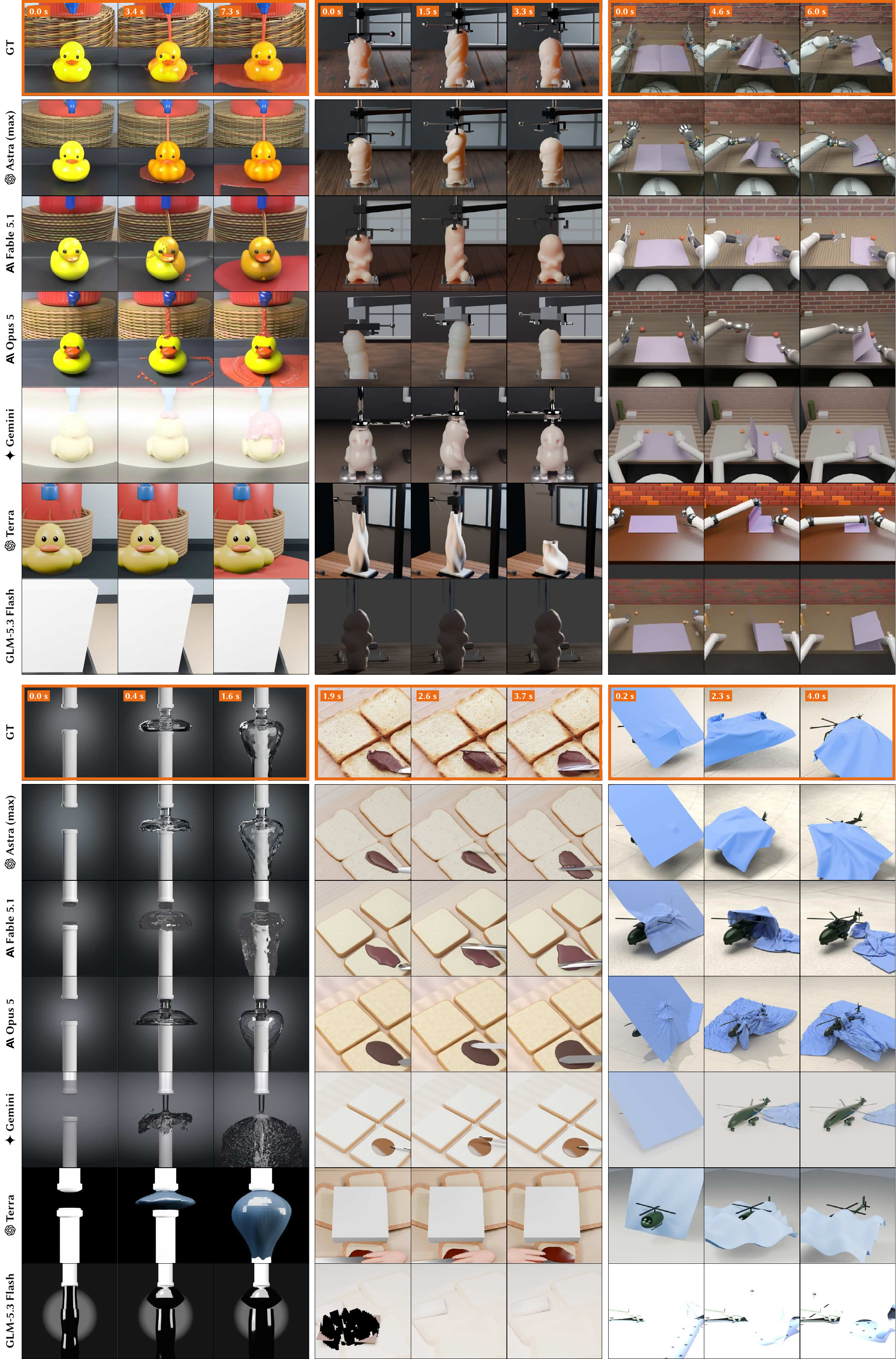}
    \caption{\textbf{Model Results.} This figure shows additional results from 6 models on 6 different \BENCHMARK tasks.}
    \label{fig:supp:results_supp_3}
\end{figure}

\subsection{VQA Results}
\label{sec:supp:results}
\begin{figure}[t]
    \centering
    \includegraphics[width=0.7\textwidth]{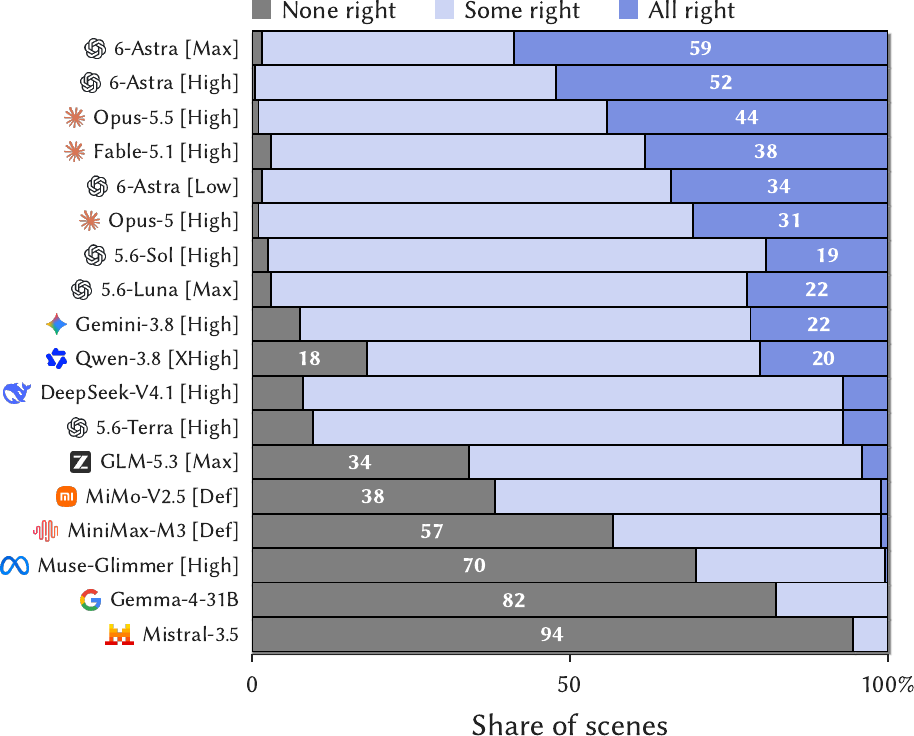}
    \caption{\textbf{Scene-level VQA.} Each bar divides a model's scenes by whether it answers none, some, or all of their questions correctly.}
    \label{fig:supp:vqa_scene_correctness}
\end{figure}

\begin{figure*}[t]
    \centering
    \begin{minipage}[b]{0.485\textwidth}
        \centering
        \includegraphics[width=\linewidth]{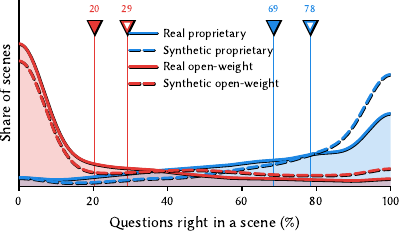}
    \end{minipage}\hfill
    \begin{minipage}[b]{0.485\textwidth}
        \centering
        \includegraphics[width=\linewidth]{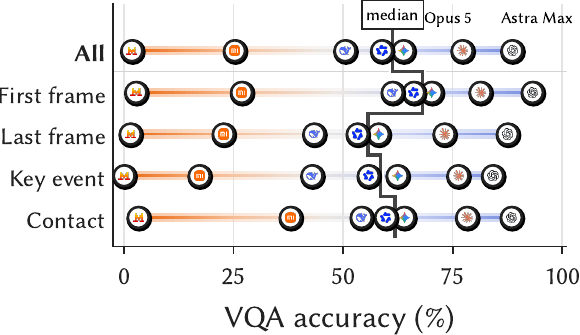}
    \end{minipage}
    \begin{minipage}[t]{0.485\textwidth}
        \caption{\textbf{VQA score distribution.} Density of the proportion of a scene's questions answered correctly over (model, scene) pairs, for proprietary and open-weight models on real and synthetic scenes; triangles mark the means.}
        \label{fig:supp:vqa_score_dist}
    \end{minipage}\hfill
    \begin{minipage}[t]{0.485\textwidth}
        \caption{\textbf{VQA by question category.} Each row spans the accuracy of all models on one category, from the worst to the best; badges mark selected models, and the stepped line connects the medians.}
        \label{fig:supp:vqa_categories}
    \end{minipage}
\end{figure*}

\begin{figure*}[ht]
    \centering
    \includegraphics[width=\textwidth]{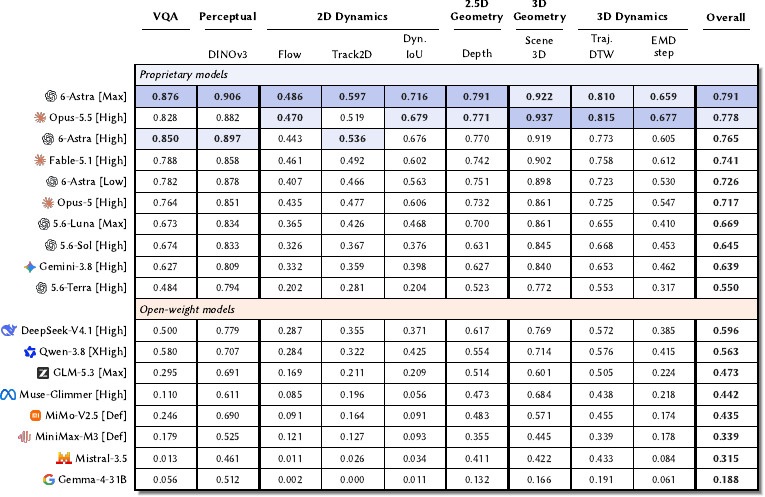}
\caption{\textbf{Expanded leaderboard.} The individual metrics behind each family score of Figure~\ref{fig:main_leaderboard}, grouped by family; higher is better, and the best and second best in each column are shaded.}
\label{fig:supp:main_metrics}
\end{figure*}

\begin{figure*}[t]
    \centering
    \includegraphics[width=\textwidth]{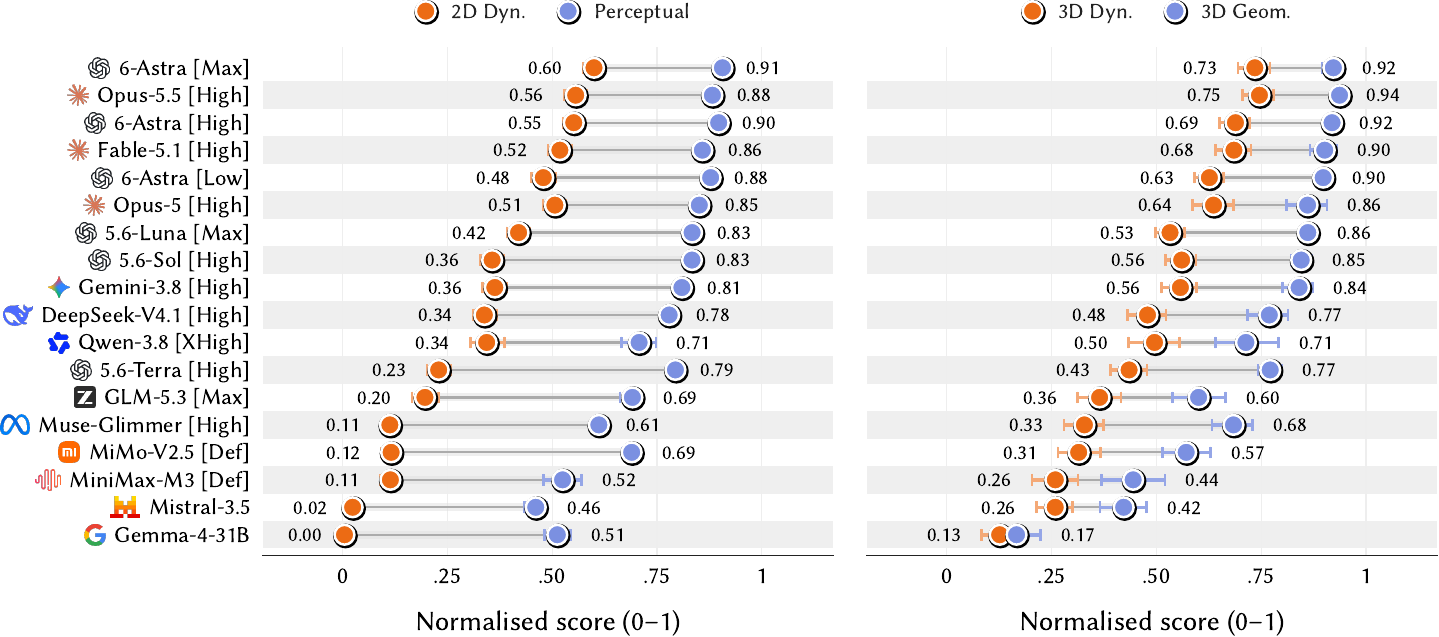}
\caption{\textbf{Static--dynamic gap by space.} The pooled gap of Figure~\ref{fig:dynamic_static_gap} separated into Perceptual against 2D Dynamics (left) and 3D Geometry against 3D Dynamics (right), with 95\% intervals over scenes.}
\label{fig:supp:family_gaps}
\end{figure*}

Figure~\ref{fig:supp:vqa_scene_correctness} splits each model's scenes by how many of their VQA questions it answers correctly. The strongest models answer every question on a majority of their scenes (Astra [Max]: 59\%), whereas the weakest open-weight models answer none on most (Mistral: 94\%). Figure~\ref{fig:supp:vqa_score_dist} shows the same split as a distribution over scenes for proprietary and open-weight models, and Figure~\ref{fig:supp:vqa_categories} breaks VQA accuracy down by question category. Figure~\ref{fig:supp:main_metrics} lists every metric behind the main leaderboard, and Figure~\ref{fig:supp:family_gaps} splits the static--dynamic gap by space.

\subsection{Comparison Between Real and Synthetic Splits}
\label{supp:real_synthetic}
\begin{figure*}[t]
    \centering
    \includegraphics[width=\textwidth]{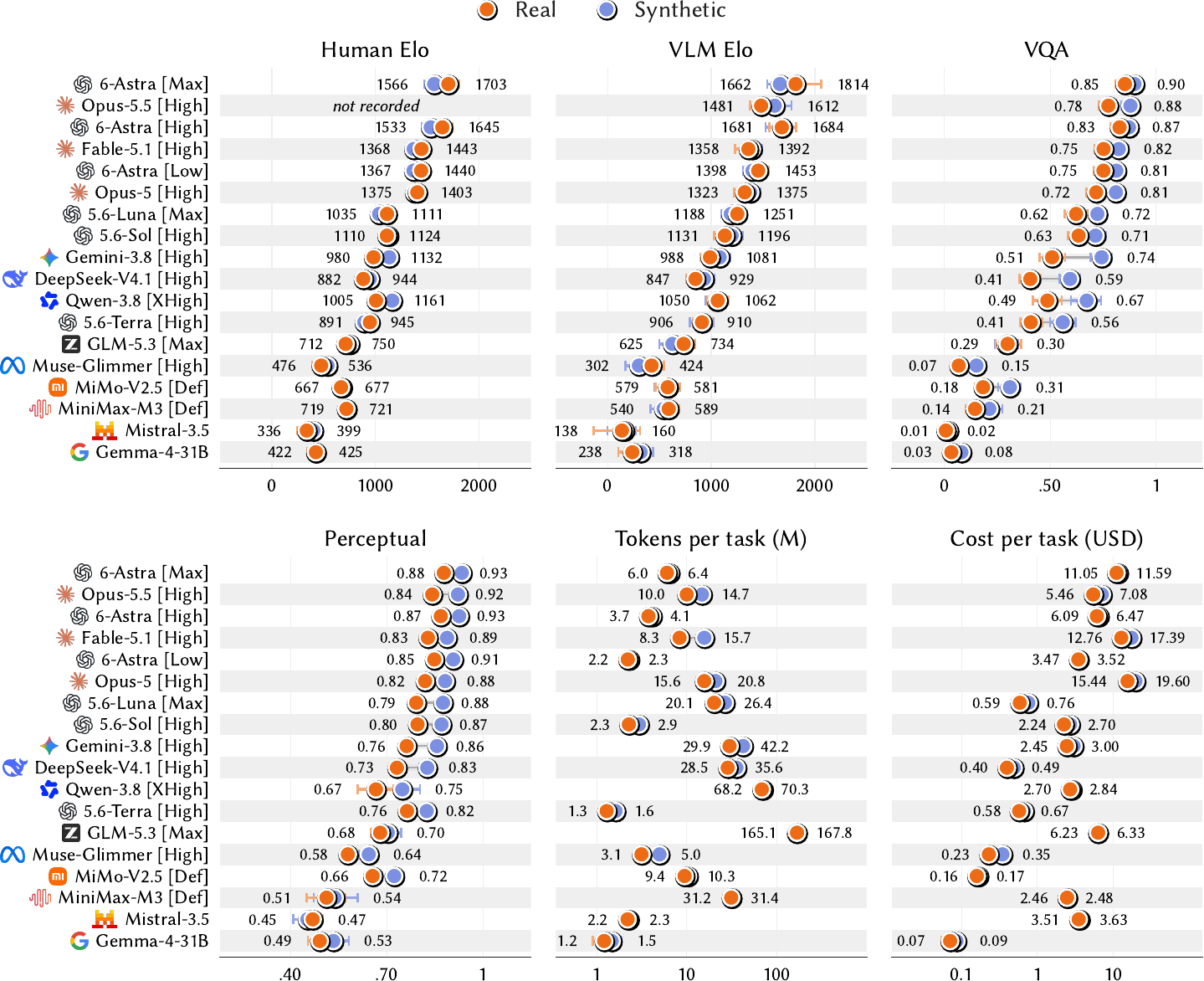}
\caption{\textbf{Real versus synthetic scenes.} For each model, the orange dot is its reading on the 100 real scenes and the blue dot on the 100 synthetic scenes, with 95\% intervals. Most models score lower on real scenes while spending fewer tokens on them, and GPT-6 Astra [Max] shows the strongest real-world generalization.}
\label{fig:supp:real_synthetic_gap}
\end{figure*}

Figure~\ref{fig:supp:real_synthetic_gap} compares each model on the 100 real and 100 synthetic scenes. Most models reconstruct real scenes worse, scoring lower on both VQA and Perceptual, and the drop is largest in the middle of the ranking (Gemini 3.8 Flash [High], DeepSeek v4.1 Flash [High], and Qwen3.8 Flash [XHigh]). The strongest models, however, differ sharply in real-world generalization. Claude Opus 5.5 [High] nearly matches GPT-6 Astra [Max] on synthetic scenes (VQA 0.88 against 0.90), but it falls well behind on real footage (0.78 against 0.85). GPT-6 Astra [Max] instead carries its performance over to real scenes and widens its Elo lead there, showing the strongest real-world generalization among all models. Agents also do not spend more effort on the harder footage, as models use 17\% fewer tokens on real scenes on average.

\subsection{How Do Models Describe Motion?}
\label{supp:describe_motion}
 Table~\ref{tab:solution-strategies} shows a breakdown of how the models attempted to solve the \BENCHMARK tasks. Blender physics means the scene is carried by Blender's own solvers, mostly point-cache bakes, the rigid-body world (105) and the cloth modifier (41), with Mantaflow used three times; custom simulations are written in the solution itself, 318 of them identifying an established method (MPM or MLS-MPM 111, PBD or XPBD 136, DEM 24, PBF 21, mass-spring 15, FEM 14, APIC 10, Verlet 9, FLIP 8) and the others integrating an unnamed scheme, which we detect from explicit time stepping (247), GPU kernels (43), neighbor queries for contact (41), impulse exchange between bodies (14) or particle-grid transfers (8); 155 run on Taichi and 102 on Warp. Analytic solutions instead evaluate a closed-form function of time. Examples include interpolating splines through chosen poses (539), a prescribed map that displaces every vertex (312), ballistic formulas with gravity written in (256), placement along a curve by arc length (236), hand-written pose tables (135) and piles rebuilt each frame to keep their volume, as a height field or a cone at a fixed slope (107).

 In three of the cases in Figure~\ref{fig:teaser}, the choice follows the phenomenon. Every model bakes the dam break with Blender's Mantaflow FLIP solver, and almost all write position-based rod simulations for the pasta strands, with Fable 5.1 and Opus 5 using PBD and XPBD with kd-tree contacts and DeepSeek v4.1 Flash writing Taichi kernels. The bread tear separates them: Fable 5.1 runs two-field MLS-MPM with a per-particle fracture threshold, DeepSeek v4.1 Flash snaps XPBD ligaments across the seam, and Astra [Max] writes no solver and instead advances a cohesive fracture front down the loaf analytically.
 
\newcommand{\modellogo}[1]{\makebox[1.45em][l]{\raisebox{-0.15em}{\includegraphics[height=0.9em,width=1.15em,keepaspectratio]{figures/logos/#1.pdf}}}}
\begin{table}[t]
\centering
\small
\caption{How models construct motion. Each solution either uses Blender physics, writes a custom simulation, computes motion analytically, or keyframes poses. Results cover the solutions of all 18 models; a few use other approaches, so some rows do not sum to 100\%. The final row pools all solutions.}
\label{tab:solution-strategies}
\begin{tabular}{lrrrr@{}}
\toprule
Model [Reasoning effort] & Blender & Custom sim. & Analytic & Keyframed \\
\midrule
\modellogo{claude}Claude Opus 5.5 [High] & 15\% & \textbf{61\%} & 22\% & 2\% \\
\modellogo{claude}Claude Opus 5 [High] & 16\% & 51\% & 25\% & 7\% \\
\modellogo{qwen}Qwen3.8 Flash [XHigh] & 6\% & 45\% & 45\% & 4\% \\
\modellogo{claude}Claude Fable 5.1 [High] & 22\% & 44\% & 27\% & 7\% \\
\modellogo{deepseek}DeepSeek v4.1 Flash [High] & 10\% & 25\% & 56\% & 7\% \\
\modellogo{gpt}GPT-6 Astra [Max] & 8\% & \textbf{24\%} &\textbf{ 63\%} & 5\% \\
\modellogo{glm}GLM-5.3 Flash [Max] & 14\% & 19\% & 56\% & 4\% \\
\modellogo{gpt}GPT-6 Astra [High] & 8\% & 14\% & 73\% & 5\% \\
\modellogo{mimo}MiMo v2.5 [Def] & 13\% & 14\% & 70\% & 3\% \\
\modellogo{gemini}Gemini 3.8 Flash [High] & 28\% & 12\% & 49\% & 11\% \\
\modellogo{mistral}Mistral Medium 3.5 & 8\% & 9\% & 83\% & 0\% \\
\modellogo{minimax}MiniMax M3 [Def] & 15\% & 8\% & 74\% & 1\% \\
\modellogo{gpt}GPT-6 Astra [Low] & 7\% & 6\% & 85\% & 3\% \\
\modellogo{gpt}GPT-5.6 Sol [High] & 5\% & 5\% & 88\% & 1\% \\
\modellogo{gpt}GPT-5.6 Luna [Max] & 0\% & 5\% & 93\% & 2\% \\
\modellogo{gemma}Gemma-4 31B & 4\% & 2\% & 94\% & 0\% \\
\modellogo{muse_glimmer}Muse Glimmer [High] & 5\% & 1\% & 92\% & 1\% \\
\modellogo{gpt}GPT-5.6 Terra [High] & 0\% & \textbf{0\%} & \textbf{100\% }& 0\% \\
\midrule
\textbf{Pooled} & 10\% & 19\% & 67\% & 3\% \\
\bottomrule
\end{tabular}
\end{table}

\clearpage
\section{Experimental Setup}
\label{sec:supp:setup}
This section details the agent execution environment (Appendix~\ref{sec:supp:harness}), the task prompt (Appendix~\ref{sec:supp:agent_prompt}), the unified 4D world submission specification (Appendix~\ref{sec:supp:output_format}), and aggregate compute statistics (Appendix~\ref{sec:supp:running_stats}).

\subsection{Execution Environment}
\label{sec:supp:harness}
To evaluate each model with its intended harness, we run proprietary models in their provider's native command-line agent: Codex CLI for GPT models, Claude Code for Claude models, and the Antigravity CLI for Gemini. We run all other models in Stirrup~\citep{stirrup}, an open agent loop driving an OpenAI-compatible endpoint with shell, image-viewing, and finish tools; these open-weight models are served via SGLang~\citep{zheng2024sglang} using recommended model configurations. Each agent maintains its native system prompt and receives the task prompt (Appendix~\ref{sec:supp:agent_prompt}) as its initial instruction.

Each run executes in a fresh, isolated container mounting the reference video (\nolinkurl{/input/reference.mp4}) and task context (\nolinkurl{/task}) read-only, alongside an empty writable workspace (\nolinkurl{/workspace}). The task directory provides the task prompt, world format specification (\texttt{world-format.md}), and render settings (\texttt{render-settings.md}). To prevent benchmark hacking and information leakage, all scene-specific metadata (e.g., scene names, descriptions, and reference 3D worlds) are stripped; operational parameters such as duration, resolution, and frame rate must be extracted directly from the video. Upon completion, only \texttt{world/} and \texttt{solution/} are collected for evaluation.

The container environment (CUDA~13.0, Ubuntu~24.04) provides Blender~5.2.0 (as both CLI and \texttt{bpy} module) with FFmpeg and ImageMagick for rendering and visual processing. For external geometry and dynamic computation, it pre-installs NumPy~2.3.1, SciPy~1.18.0, PyTorch~2.13.0, Taichi~1.7.4, Warp~1.17.0, Trimesh~5.0.0, and Pillow~11.3.0. It also includes an offline copy of the Blender~5.2 API reference and the public format checker (\texttt{python -m checker}), but not the evaluation scorer. We impose no artificial step limits; only runs that stall or exceed a generous 6\,h timeout are terminated.

\subsection{Agent Prompt}
\label{sec:supp:agent_prompt}

\begin{tcolorbox}[enhanced, breakable, colback=gray!4, colframe=gray!55, boxrule=0.5pt, arc=2pt, left=5pt, right=5pt, top=4pt, bottom=4pt,
    title={Task prompt (\texttt{TASK.md})}, coltitle=black, colbacktitle=gray!15, fonttitle=\footnotesize\bfseries]
\begin{Verbatim}[fontsize=\scriptsize, breaklines]
`input.mp4` is a video of a dynamic scene.
Your task is to write graphics code that, when rendered, reproduces the observed 4D scene.

Your reconstruction should model the scene and its dynamics explicitly.
It should be both visually and physically consistent with the target.
Render the result in Blender, matching the input's dimensions, framerate, and camera.

## Scoring
Your solution will be scored frame-by-frame.
It will also be evaluated against the ground-truth 3D geometry.
Accordingly, solutions that represent motion or appearance in 2D or copy from the video instead of explicitly modeling the scene will be strongly penalized.
The autograder will fail submissions that copy pixels from the input video into the Blender scene for any purpose, such as to texture objects; all must be programmatically defined.

## Output
The output should contain two directories:
1. `/workspace/solution` is the program, an executable `build.sh`, and the Python it runs.
2. `/workspace/world` is its output, holding the camera, the render, the geometry at every frame, and the 3D state of the moving matter over time.

`build.sh` takes no arguments, does **not** read the video file, and writes all of `/workspace/world` from nothing.
Both are required to verify that the reconstruction is reproducible.

Run `python -m checker` to confirm that what you have written is in the correct format.
The packaged checker does not contain the privileged autograder.

## How to build the scene and its motion

One option is to do all of it in Blender: describe the 3D scene and its motion
through time there, and render that 4D sequence. You can also write external
graphics code or programs using the libraries provided to describe the motion of the scene
through time, and use Blender only to render the final 4D sequence. NumPy,
SciPy, PyTorch, Taichi, Warp and Trimesh are installed, and anything else on the
machine is fair game.

However the motion is computed, the evolution of the scene has to be rendered
with Blender, through the camera you export.

The rendered geometry in `meshes/` may change freely from frame to frame, vertex count and
topology included, as happens with liquids or fractures. `dynamics/` is where the
frame-to-frame correspondence lives: column `j` of `pos` is the same piece of
material at every frame, so `pos[t, j]` and `pos[t+1, j]` are that matter one
frame apart. Geometry rebuilt independently at each frame cannot give you that.

## Output format

The full format of `/workspace/world` is in `/task/world-format.md`.
It lists every file with its dtype and shape, the camera conventions, and what to return for a liquid.
Read it before you write your export.

### Program: `/workspace/solution`

`/workspace/solution` should contain an executable `build.sh` and the `.py` and `.sh` files it runs, in any layout.
Running `bash build.sh` should write the complete `/workspace/world`, replacing anything already there.
It cannot read the video, access the network, install packages, or load assets shipped alongside it.
Anyone with this environment should be able to rebuild your world by running it.

### World: `/workspace/world`

```
world/
  camera.json          {"intrinsics": K, "extrinsic": E}, the fixed camera
  render.mp4           the rendered video
  meshes/<id>.npz      each rendered object's triangles at every frame
  dynamics/<name>.npz  each piece of moving material's positions over time
  solver/              a liquid solver's own bake, if it keeps no particle identity
```

`meshes/` is the geometry that gets rendered, and its vertex count and topology may change from frame to frame.
`dynamics/` describes how the 3D geometry of the moving objects moves through time: `pos` has shape `(F, N, 3)`, and column `j` is the same piece of material at every frame.
Static objects need only a `meshes/` archive.
The points in `dynamics/` are the vertices of your simulation mesh for a rigid body, cloth or soft body, and the particles for MPM, SPH or sand.

All positions are in world-space metres, with Z up.
`render.mp4` must match `/input/reference.mp4` in frame count, frame rate and resolution, or it will not be scored.
Every file in `meshes/` and `dynamics/` uses the same `F` frames.
Each frame of `render.mp4` should show exactly the triangles in `meshes/` and nothing else.

### Environment

The environment has FFmpeg, FFprobe, ImageMagick, Python and Blender.
You can run Blender as the `blender` executable or import it in Python as `bpy`.
NumPy, SciPy, PyTorch, Taichi, Warp, Trimesh, Pillow and the checker are available in Python.
The Blender Python API reference is in `/opt/blender-docs`.
`/task/render-settings.md` has the render settings we expect.

### Checking your work

`python -m checker` reports every format error, and a world that fails it scores zero.
`python -m checker.preview` saves `/workspace/tmp/preview_0000.png`, with frame 0 of `render.mp4` next to your `meshes/` projected through `camera.json`.
If the two don't line up, your camera or meshes don't match what you rendered.
Neither tool checks whether your world matches the video.
Put intermediate files in `/workspace/tmp`, not in `world/` or `solution/`.
\end{Verbatim}
\end{tcolorbox}

\subsection{Submission Format}
\label{sec:supp:output_format}
To enable automated, unified evaluation directly from the true 4D state rather than relying solely on 2D video appearance, we require models to submit explicit 4D assets beyond the rendered video. Specifically, each submission must provide an executable \texttt{build.sh} inside \texttt{solution/} that deterministically regenerates the explicit \texttt{world/} representation, ensuring full reproducibility. Crucially, this export specification is designed to be lightweight, typically requiring only about 100 lines of standard Python export code. 
In practice, format compliance poses little barrier for most models: among runs that pass the initial video gate, only 5.4\% fail subsequent 4D world validation ($\leq 2.5\%$ across proprietary models and 3.5--7.5\% across open-weight models). The one exception is Gemma-4 31B (Figure~\ref{fig:executability}). For all other models, benchmark failures thus stem from 4D reasoning and simulation limits rather than formatting friction.

A submission consists of two directories: \texttt{solution/} contains the standalone program (\texttt{build.sh} and accompanying scripts) that deterministically regenerates the reconstruction without network access or external assets, while \texttt{world/} stores the output artifacts:
\begin{Verbatim}[fontsize=\scriptsize]
world/
  camera.json          {"intrinsics": K, "extrinsic": E}
  render.mp4           the rendered video
  meshes/<id>.npz      vertices_TTTT (V_t,3) float32, faces_TTTT (T_t,3) int32
  dynamics/<name>.npz  pos (F,N,3) float32, ids (M,) uint16; faces (T,3) or tets (K,4) int32
  solver/              the solver's bake, for an identity-less liquid only
\end{Verbatim}
All positions are in world-space meters ($+z$ up), and all frame-indexed data strictly align with the $F$ frames of \texttt{render.mp4}.

\paragraph{Camera.} \texttt{camera.json} defines a fixed pinhole camera shared across frames: a $3\times3$ intrinsic matrix $K$ (principal point at image center) and a $4\times4$ camera-to-world extrinsic transform $E$ ($+x$ right, $+y$ down, $+z$ forward).

\paragraph{Meshes.} \texttt{meshes/} contains one archive per rendered object, storing vertices and triangle indices per frame. Vertex count and topology may vary arbitrarily across time; objects absent from \texttt{dynamics/} are treated as static.

\paragraph{Dynamics.} \texttt{dynamics/} tracks moving matter across the $F$ frames. Column $j$ of \texttt{pos} follows a persistent physical point, taking finite values over its lifespan and NaN otherwise. Meshes provide surface (\texttt{faces}) or tetrahedral (\texttt{tets}) connectivity for rigid/deformable bodies, or omit connectivity for particle systems (MPM, SPH, granular).

\paragraph{Liquids.} Particle-based fluids export directly via \texttt{dynamics/}. For Eulerian solvers without persistent particle identity (e.g., Blender Mantaflow), raw simulation bakes are retained under \texttt{solver/} alongside a \texttt{manifest.json} (specifying domain coordinates, resolution, and time scale), from which particle trajectories are extracted at scoring time.

\paragraph{Render.} \texttt{render.mp4} is an H.264 video rendered in Blender through \texttt{camera.json}. Each frame must display exclusively the geometry in \texttt{meshes/} (lighting and materials are left unconstrained), without copying pixels from the reference video.

\paragraph{Checks.} To help agents catch low-level formatting or syntax errors during development and avoid accidental execution failures, we provide a public offline checker (\texttt{python -m checker}). At scoring time, the video gate enforces valid decoding and resolution and frame-rate parity, while the world gate verifies the presence and structural integrity of \texttt{camera.json}, \texttt{meshes/}, \texttt{dynamics/}, and \texttt{build.sh}.

\subsection{Running Statistics}
\label{sec:supp:running_stats}
\begin{figure*}[t]
    \centering
    \includegraphics[width=\textwidth]{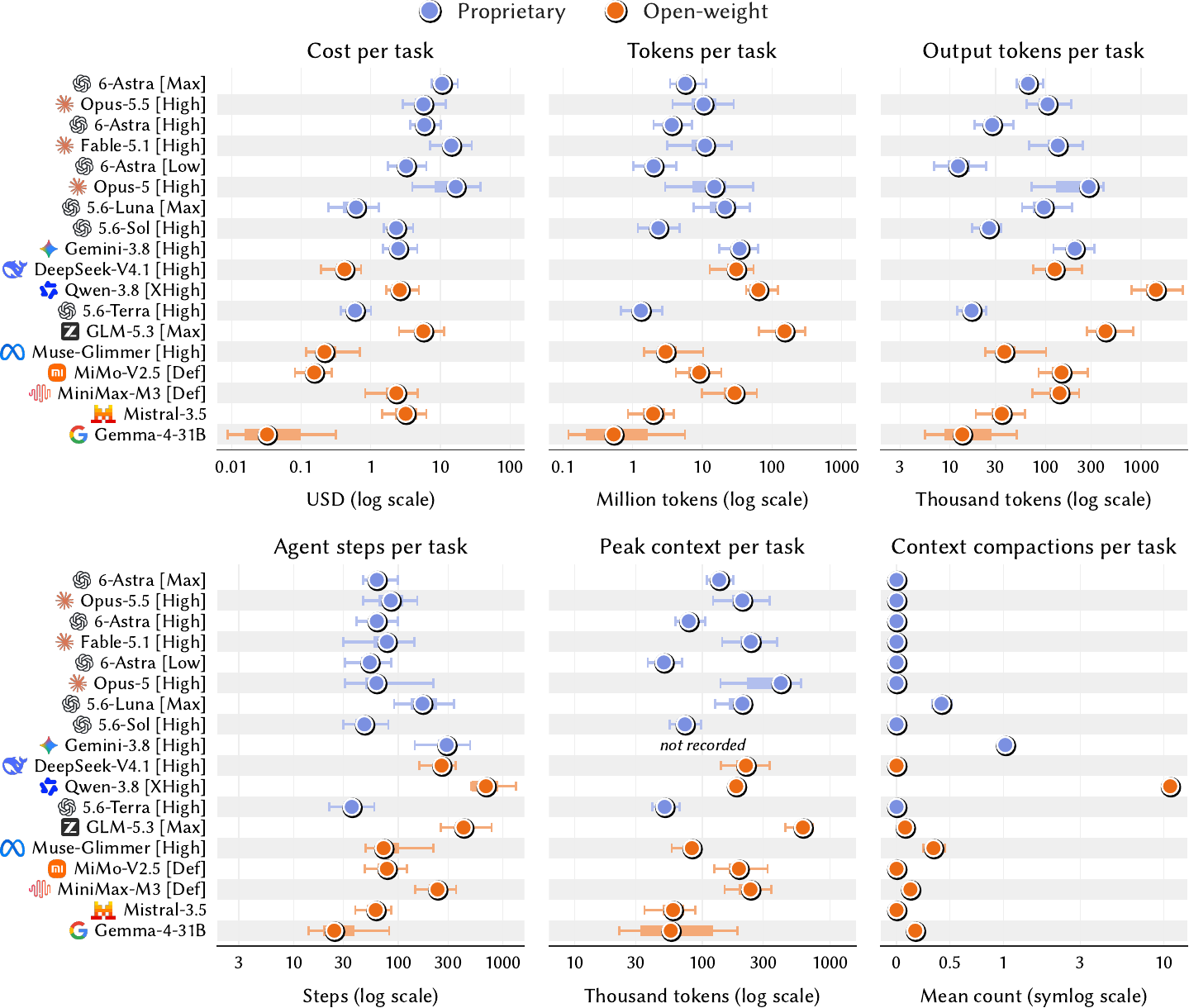}
\caption{\textbf{Running statistics.} The first five panels show, over each model's runs, the median (dot), interquartile range (thick bar), and 5th--95th percentile range (whisker) of the cost, tokens, output tokens, agent steps, and peak context per task; the last panel shows the mean number of context compactions per task with its bootstrap 95\% interval. The Antigravity CLI does not record peak context.}
\label{fig:supp:running_stats}
\end{figure*}
\needspace{24\baselineskip}
\begin{wraptable}{r}{0.48\textwidth}
\vspace{-1em}
\centering
\footnotesize
\setlength{\tabcolsep}{2.5pt}
\renewcommand{\arraystretch}{0.92}
\caption{\textbf{Token breakdown.} Mean tokens per task (M), by billing category.}
\label{tab:supp:token_breakdown}
\begin{tabular}{@{}lrrr@{}}
\toprule
Model & \begin{tabular}[b]{@{}r@{}}Cache\\write\end{tabular} & \begin{tabular}[b]{@{}r@{}}Cache\\read\end{tabular} & Output \\
\midrule
\modellogo{gpt}Astra [Max] & 0.20 & 5.96 & 0.07 \\
\modellogo{claude}Opus 5.5 [High] & 0.20 & 12.05 & 0.11 \\
\modellogo{gpt}Astra [High] & 0.10 & 3.79 & 0.03 \\
\modellogo{claude}Fable 5.1 [High] & 0.26 & 11.58 & 0.14 \\
\modellogo{gpt}Astra [Low] & 0.06 & 2.17 & 0.01 \\
\modellogo{claude}Opus 5 [High] & 0.38 & 17.57 & 0.25 \\
\modellogo{gpt}Luna [Max] & 0.46 & 22.72 & 0.11 \\
\modellogo{gpt}Sol [High] & 0.09 & 2.49 & 0.03 \\
\modellogo{gemini}Gemini 3.8 [High] & 2.94 & 32.83 & 0.21 \\
\modellogo{deepseek}DeepSeek v4.1 [High] & 0.22 & 31.69 & 0.13 \\
\modellogo{qwen}Qwen3.8 [XHigh] & 2.04 & 65.70 & 1.51 \\
\modellogo{gpt}Terra [High] & 0.07 & 1.36 & 0.02 \\
\modellogo{glm}GLM-5.3 [Max] & 0.67 & 165.33 & 0.45 \\
\modellogo{muse_glimmer}Muse [High] & 0.10 & 3.90 & 0.05 \\
\modellogo{mimo}MiMo v2.5 [Def] & 0.20 & 9.52 & 0.15 \\
\modellogo{minimax}MiniMax M3 [Def] & 0.25 & 30.88 & 0.14 \\
\modellogo{mistral}Mistral 3.5 & 0.06 & 2.13 & 0.04 \\
\modellogo{gemma}Gemma-4 31B & 0.09 & 1.22 & 0.02 \\
\bottomrule
\end{tabular}
\vspace{-2em}
\end{wraptable}
To quantify the computational and economic footprint of agentic 4D reconstruction, Figure~\ref{fig:supp:running_stats} reports per-task execution statistics across all 3,600 runs, totaling 5,983 hours and \$16,495 in compute (open-weight models priced at OpenRouter rates). Since each agent step re-sends its context during multiple render-and-compare rounds, 97\% of all tokens are cache reads (Table~\ref{tab:supp:token_breakdown}). Cache write counts all uncached input, and output includes reasoning. For open-weight models, the cache hit rate depends on the serving setup, so we assume perfect prefix caching and a 95\% cache hit rate when computing cost. Per task, open-weight models expend a mean of 247 agent steps and 39.6M tokens compared to 101 steps and 11.8M tokens for proprietary models, yet cost less (mean \$2.01 vs.\ \$6.65) due to lower token pricing.

Crucially, sheer agent loop activity does not inherently yield better 4D reconstructions across models: while per-task cost correlates moderately with the Overall score (Spearman $\rho=0.65$), raw step counts and token volume exhibit negligible correlation ($|\rho|\le0.10$). For example, Qwen [XHigh] expends an extreme mean of 752 steps and 1.5M output tokens, yet ranks eleventh among 18 models. Conversely, holding model architecture fixed, increasing test-time reasoning scales performance monotonically: raising GPT-6 Astra from [Low] to [High] to [Max] increases output tokens (13K $\rightarrow$ 29K $\rightarrow$ 67K) and cost (\$3.5 at [Low] to \$11.3 at [Max]), steadily improving its Overall score from 0.726 to 0.791.

\end{document}